\pdfoutput=1
\documentclass[journal]{IEEEtran}

\usepackage{amsmath,amsfonts}
\usepackage[ruled,vlined,linesnumbered]{algorithm2e}
\usepackage{array}
\usepackage{subcaption}
\usepackage{textcomp}
\usepackage{stfloats}
\usepackage{url}
\usepackage{verbatim}
\usepackage{graphicx}
\usepackage{cite}
\usepackage{multirow}
\usepackage{lipsum}
\usepackage{mathtools}
\usepackage{cuted}
\usepackage{breqn}
\usepackage[english]{babel}
\usepackage[autostyle]{csquotes}
\usepackage[acronym,toc,shortcuts]{glossaries}
\usepackage{comment}
\usepackage{siunitx}
\usepackage{textgreek}
\usepackage{xcolor}

\begin{document}

\title{Hierarchical Federated Learning in Multi-hop Cluster-Based VANETs}

\author{M. Saeid HaghighiFard,~\IEEEmembership{Student Member,~IEEE} and Sinem Coleri,~\IEEEmembership{Fellow,~IEEE}
		
\thanks{M. Saeid Haghighifard and Sinem Coleri are with the Department of Electrical and Electronics Engineering, Koc University, Istanbul, Turkey, email: {mhaghighifard21, scoleri}@ku.edu.tr. This work is supported by the Scientific and Technological Research Council of Turkey Grant number 119C058 and Ford Otosan.}
}

\markboth{Journal of \LaTeX\ Class Files,~Vol.~14, No.~8, August~2021}%
{Shell \MakeLowercase{\textit{et al.}}: A Sample Article Using IEEEtran.cls for IEEE Journals}

 \maketitle

 \IEEEtitleabstractindextext{%
    \begin{abstract}
        The usage of federated learning (FL) in Vehicular Ad hoc Networks (VANET) has garnered significant interest in research due to the advantages of reducing transmission overhead and protecting user privacy by communicating local dataset gradients instead of raw data. However, implementing FL in VANETs faces challenges, including limited communication resources, high vehicle mobility, and the statistical diversity of data distributions. In order to tackle these issues, this paper introduces a novel framework for hierarchical federated learning (HFL) over multi-hop clustering-based VANET. The proposed method utilizes a weighted combination of the average relative speed and cosine similarity of FL model parameters as a clustering metric to consider both data diversity and high vehicle mobility. This metric ensures convergence with minimum changes in cluster heads while tackling the complexities associated with non-independent and identically distributed (non-IID) data scenarios. Additionally, the framework includes a novel mechanism to manage seamless transitions of cluster heads (CHs), followed by transferring the most recent FL model parameter to the designated CH. Furthermore, the proposed approach considers the option of merging CHs, aiming to reduce their count and, consequently, mitigate associated overhead. Through extensive simulations, the proposed hierarchical federated learning over clustered VANET has been demonstrated to improve accuracy and convergence time significantly while maintaining an acceptable level of packet overhead compared to previously proposed clustering algorithms and non-clustered VANET.
    \end{abstract}
}

\maketitle
\IEEEdisplaynontitleabstractindextext


\section{Introduction}

\IEEEPARstart{V}{ehicular} Ad hoc NETworks (VANETs) have recently adopted machine learning (ML) algorithms to improve the safety of transportation networks by providing a set of tools to assist the system in making more informed and data-driven decisions based on learning from sensors such as LIDAR, RADAR, vehicle cameras, and provide novel services such as location-based services, real-time traffic flow prediction and control, and autonomous driving~\cite{5395779}. In centralized machine learning algorithms, a robust learning algorithm, typically a neural network (NN), is trained on a vast dataset acquired from the vehicles' edge devices~\cite{elbir2022federated}. The concept of federated learning (FL) was recently established to bring machine learning (ML) to the edge~\cite{9084352}, which reduces transmission overhead and preserves the privacy of the users. In FL, instead of sending local datasets to the central entity, the edge devices only communicate the gradients of the learnable parameters obtained from these local datasets. These gradients are aggregated by the central entity, which determines the model parameters, which are subsequently sent to the edge devices. This process is repeated until the learning model is fully trained~\cite{elbir2022federated}. The main challenges of using FL algorithms are restricted communication resources of the wireless environment, vehicle mobility, and statistical heterogeneity of data distributions over the road.

The first challenge of the restricted communication resources in FL arises when the central entity cannot satisfy the concurrent requests from a massive number of vehicles, resulting in a communication bottleneck. Since the produced traffic grows linearly with the number of participating vehicles and model size, one way to address the communication bottleneck problem is a partial participation rule to reduce the number of vehicles uploading updates simultaneously~\cite{9505307}. However, the convergence time of the FL may significantly increase with partial participation, which may not be acceptable for some applications~\cite{9716076}. Another way to alleviate the bottleneck problem is the usage of hierarchical FL such that clusters of vehicles train models concurrently and only the aggregate of each cluster is provided by the cluster head to the central entity~\cite{9054634,jabri2019vehicular,8437169}. Cluster formation in FL has been formulated as an optimization problem with different objective functions and constraints~\cite{9716076,9174890,9831009}.~\cite{9716076} performs the clustering with the objective of maximizing the weighted sum of the FL accuracy of the cluster heads, with the constraint that the sum of training and upload time of any vehicle is less than the link lifetime, which is the duration of time where two vehicles remain connected.~\cite{9174890} formulates the clustering problem with the objective of minimizing the cosine similarity of the client gradient updates while constraining on the maximum similarity between the updates from any two clients based on their data generating distribution.~\cite{9831009} aims to minimize the empirical risk of the model over the data samples and parameters of the FL model while constraining the data samples in each cluster should not overlap with those of the other clusters. However, these clustering algorithms are impractical since they require collecting complete network topology and data characteristics of the nodes in a highly dynamic vehicular network.

\begin{figure*}[t]
\setlength\belowcaptionskip{0pt}
\centering
\includegraphics[width=15 cm]{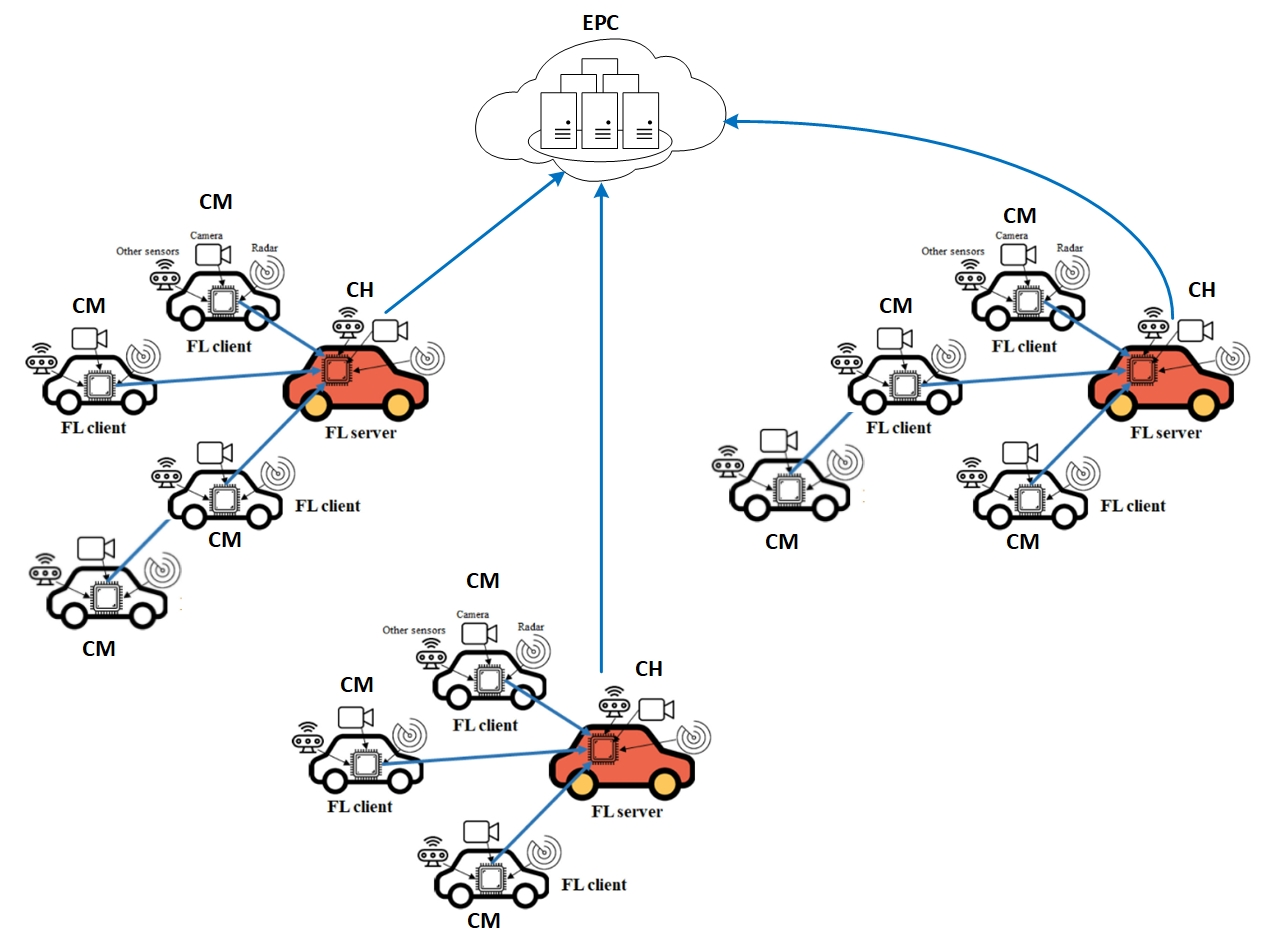}
\caption{Hierarchical FL over Multi-hop Clustering in VANET}
\end{figure*}

The second challenge of handling high vehicle mobility in FL arises since network coverage, link quality, and stability fluctuate rapidly in vehicle-to-everything (V2X) communication, which influences the performance of FL applications regarding convergence rate and accuracy~\cite{elbir2022federated}.~\cite{9716076} considers the mobility in clustered FL by constraining that the upload time for the transmission of the model update to the cluster head is less than the time duration during which the vehicle is under the coverage of the cluster head or central entity.~\cite{9831009} introduces route vehicles as a relay in clustered FL in the case that cluster members are out of the coverage of the cluster head. Route vehicles connect to the cluster head and a member at the same time, computing model parameters depending on the parameters provided by the member and forwarding the new aggregated model to the cluster head. However, all of these papers assume that FL learning time is shorter than cluster formation duration in their optimization problem, and no algorithm has been proposed to deal with clusters changing faster than the convergence time of the FL learning.

The third challenge of handling the heterogeneity of data distributions in FL arises since the non-IID data of the nodes in the network cause slow convergence of the FL model~\cite{9054634,9220170} and reduces model accuracy significantly when compared to centralized learning~\cite{mcmahan2017communication}.~\cite{9831009} proposes a weighted update for inner-cluster FL algorithm to give greater weight to the cluster with fewer data samples. This improves the accuracy and convergence time of the FL by incorporating the data of each vehicle into the model in a fair manner.~\cite{ghosh2020efficient} and~\cite{9207469} provide a clustering algorithm for FL based on the similarity of their local updates such that the data of the nodes in each cluster are IID. On the other hand,~\cite{9716076} proposes a cluster-head selection algorithm, where the node with the highest similarity of the local updates to the remaining cluster members is selected as cluster head. However, none of these papers address the slow convergence problem of non-IID data in HFL while considering the high mobility of vehicles.

This paper proposes a novel framework for hierarchical federated learning (HFL) in clustered VANETs based on the usage of a practical clustering algorithm and metrics that take into account both the mobility of the vehicles and non-IID data structure while introducing a mechanism to adapt FL to CH changes during the execution of the FL. Vehicles establish multi-hop clusters where cluster members (CMs) transmit their local gradients to the CHs. These CHs then propagate the average computed model parameter back to the cluster members and the EPC. The EPC computes the global model parameters and relays them to the CHs, thereby updating the entire system model. The multi-hop distributed clustering algorithm is employed to maximize clustering stability and minimize the convergence time of FL with minimum overhead. The proposed cluster head selection metric integrates average relative speed and cosine similarity for efficient federated learning with rapid convergence. Moreover, a mechanism to adapt FL to the CH changes in the network is incorporated by transferring the most recent learning parameters to the new CH for use as an edge node aggregator. The original contributions of the paper are listed as follows:

\begin{itemize}
\item We propose a novel framework for the implementation of FL in vehicular networks while simultaneously considering restricted communication resources, vehicle mobility, and non-IID data structure for the first time in the literature.
\item We propose a novel dynamic distributed multi-hop clustering algorithm for the efficient implementation and rapid convergence of the HFL for the first time in the literature. The clustering metric integrates vehicle mobility, represented by average relative speed, and data similarity, as assessed by cosine similarity. Average relative speed is vital for handling dynamic vehicular movements, ensuring cluster stability. Cosine similarity ensures accurate model parameter alignment among cluster members, facilitating rapid convergence in HFL.
\item We propose a novel hierarchical FL algorithm over the multi-hop clustered architecture to address the bottleneck issue due to restricted communication resources by combining distributed learning and computing for the first time in the literature. The HFL algorithm is capable of dealing with the divergence possibility in non-IID datasets and the high mobility of vehicles by employing intelligent clustering metrics and adapting to the cluster head changes during the execution of the FL.
\item We demonstrate via extensive simulations that our proposed methodology outperforms previously proposed non-clustered and cluster-based VANET regarding convergence time, adaptation to data heterogeneity, and communication overhead.
\end{itemize}
The subsequent sections of this paper are structured as follows: In Section II, the system model is detailed. Section III introduces the proposed multi-hop FL-based clustering algorithm. Section IV provides the hierarchical FL over the clustered VANET algorithm. Section V presents the performance evaluation of the proposed HFL over clustered VANET algorithm compared to previously proposed algorithms for different network sizes and transmission range values. Finally, Section VI presents concluding remarks and future research directions.

\section{System Model}
We assume a hierarchical multi-hop clustered structure in each direction of the road. The network contains cluster heads (CHs), cluster members (CMs) that connect to CHs in multiple hops, and the evolving packet core (EPC) as the central entity that communicates with all the CHs. Vehicles contain two communication interfaces. The first interface is for vehicle-to-vehicle (V2V) communications by using such standards as IEEE 802.11p~\cite{5888501} and IEEE 802.11bd for Wireless Access in Vehicular Environments (WAVE)~\cite{9779322}, standard cellular technologies, including LTE D2D~\cite{7497762} and 5G NR V2X~\cite{9392787}. The second interface enables vehicle-to-infrastructure (V2I) or with cloud networks (V2N) communications by using such standards as LTE V2X~\cite{7497762} and 5G NR V2X~\cite{9392787}.

The vehicles are organized into clusters in a distributed manner due to their highly dynamic behavior, and their role as CH and CM may change over time. The metrics used in this clustering are average relative speed and similarity of their local model parameters.

The hierarchical FL algorithm is executed over the hierarchical multi-hop clustered VANET structure, as shown in Fig. 1. CMs act as clients that provide model parameter updates to the CHs they are associated with based on their own local data, which are gathered in a non-IID distribution structure. CHs correspond to edge nodes that aggregate the model parameter updates collected from the CMs and send the latest updates to the EPC periodically for the general model aggregation. EPC acts as the central entity that collects and combines the aggregated model updates from the CHs. The EPC then forwards the latest updates to the CHs, which transmit them to the CMs within their clusters. The model updates are calculated by each CM using stochastic gradient descent (SGD). 

Each vehicle contains a vehicle information base (VIB) that stores information on the vehicle and its nearby vehicles within a predefined maximum number of hops. The vehicle information includes the node's direction, location, velocity, current clustering state, the number of hops to the CH if it is a CM, the ID of the vehicle that connects the node to the cluster, the ID of the vehicles that use the node to connect to the CH, the clustering metric, the model parameters, number of data points and the cosine similarity of the local parameter updates. VIB is updated whenever the vehicle's own information changes or when it receives a periodic $HELLO\_PACKET$ or request for model parameters from any of the neighbors~\cite{7081788}. Each node must have enough storage capacity and computing power to store and calculate model parameters in FL.

\section{Multi-hop FL-Based Clustering algorithm}
We propose a distributed multi-hop clustering algorithm for the implementation of hierarchical FL. In addition to the commonly used relative mobility measure in previous clustering algorithms designed for various applications~\cite{7081788}, we use the similarity between the latest learning model parameters in FL. The relative mobility measure is a function of the average relative speed with respect to the nearby vehicles. The incorporation of relative mobility into the clustering metric aims to maximize the stability of the clusters with minimum change in CH and CM roles in the clusters. On the other hand, the similarity between the latest learning model parameters is used to mitigate the non-IID issue in FL, which causes a slow convergence in the global model learning. The proposed clustering algorithm reduces the clustering overhead by exchanging the relative speed and FL model parameters only with the neighboring nodes, establishing a direct link to a neighbor who is already a head or a member of a cluster rather than connecting to the CH over multiple hops and focusing on clustering based on both mobility and model update similarity metric to preserve the cluster structure while mitigating the probability of divergence in the FL algorithm.

At any particular moment, each vehicle is in one of the five states listed below~\cite{7081788}.
\begin{itemize}
\item  INITIAL (IN): Starting state of the vehicle.
\item  STATE ELECTION (SE): Vehicle state in which it decides on the next state based on the information in VIB.
\item CLUSTER HEAD (CH): Vehicle state in which it is stated to be a cluster head and is connected to an EPC.
\item CLUSTER MEMBER (CM): Vehicle state in which the vehicle is connected to an existing cluster head.
\end{itemize} 
The vehicle begins in its initial state and remains there for an $IN\_TIMER$ duration, as shown in Algorithm 1. The reference time unit is denoted as $time$, and each communication round is considered as a single FL iteration within the FL procedure, denoted by $i$. The vehicle builds its own $VIB$ by periodically exchanging $HELLO\_PACKET$s with its neighboring nodes (Lines 1-2). $HELLO\_PACKET$ contains vehicle information, including its direction, location, velocity, current clustering status, number of hops to the CH if it is a CM, the ID of the vehicle that connects the node to the cluster, cosine similarity of FL model parameters, average relative speed with respect to all neighboring nodes, and number of local samples for each vehicle. After $IN\_TIMER$ duration, the vehicle proceeds to state $SE$ (Line 3).

\begin{algorithm}
\While{$time$ $<$ $IN\_TIMER$}{
    Update VIB upon reception of $HELLO\_PACKET$\;
    }
    $STATE = SE$\;   
\caption{$IN$ State in Multi-hop FL-based Clustering in
VANETs}
\end{algorithm}

\begin{algorithm}
\For { all $CH \in VIB$}{
$CONNECTION_{CH} = false$
}
\For { all $CH \in VIB$ in one hop}
   {
   \If{$CONNECTION_{CH} = false$}
    {
    Send $JOIN\_REQUEST$\;
        \If{$JOIN\_RESPONSE$ received}
            {
            $Avg\_CoSim_{CH} = (\alpha (|Speed_{CH}-Speed_{k}|) + (1 - \alpha)(1 - Cosine\_Similarity (\theta_{CH,i},\theta_{k,i})))$\;
            Add $Avg\_CoSim_{CH}$ to $VIB$ for $CH$\;
            $CONNECTION_{CH} = True$\;
            } 
        }
    }
\While {$\exists CH \in VIB$ with $CONNECTION_{CH} = True$}
    {
        $CH_{k}= \underset{CH \in VIB}{\operatorname{argmin}} (Avg\_CoSim_{CH})$\;
        Send $CONNECT\_REQUEST$ to $CH_k$\;
        \If {$CONNECT\_RESPONSE$ received}
        {
	$STATE = CM$\;
        Exit\;
        }
        \Else{
        $CONNECTION_{CH_k} = False$\;
        }
    }
\For { all $CM \in VIB$}{
$CONNECTION_{CM} = false$
}
\For {all $CM \in VIB$}
    {
    \If{$CONNECTION_{CM} = false$ and Number of hops $<$ $MAX\_HOP-1$}
            {
            Send $JOIN\_REQUEST$\;
            \If{$JOIN\_RESPONSE$ received}
            {
                {
                $Avg\_CoSim_{CM} = (\alpha (|Speed_{CM}-Speed_{k}|) + (1 - \alpha)(1 - Cosine\_Similarity (\theta_{CM,i},\theta_{k,i})))$\;
                Add $Avg\_CoSim_{CM}$ to $VIB$ for $CM$\;
                $CONNECTION_{CM} = True$\;
                }
           }
           }
    }
\While {$\exists CM \in VIB$ with $CONNECTION_{CM} = True$}
    {
        $CM_{k}= \underset{CM \in VIB}{\operatorname{argmin}} (Avg\_CoSim_{CM})$\;
        Send $CONNECT\_REQUEST$ to $CM_k$\;
        \If {$CONNECT\_RESPONSE$ received}
        {
	$STATE = CM$\;
        Exit\;
        }
        \Else{
        $CONNECTION_{CM_k} = False$\;
        }
    }
\For { all $SE \in VIB$}{
    $Avg\_CoSim_{SE} = (\alpha (Average\_Speed_{SE}) + (1 - \alpha)(1 - Cosine\_Similarity (\theta_{SE,i},\theta_{SE,i-1})))$\;
    Add $Avg\_CoSim_{SE}$ to $VIB$ for $SE$\;
    }
\If{$Avg\_CoSim_{k} < min_{SE \in VIB} (Avg\_CoSim_{SE})$}
    {
    $STATE = CH$\;
    }
        
\caption{$SE$ State in Multi-hop FL-based Clustering in
VANETs}
\end{algorithm}

In state $SE$, the vehicle should decide whether to join to a CH, join to a CM, become the CH, or stay in $SE$, as shown in Algorithm 2. The vehicle should first determine if it can connect to a CH or a CM within one hop. The vehicle first tries to connect to a CH within one hop. To do this, it sets the connection to all of these CHs to false (Lines 1-2). The vehicle then sends the $JOIN\_REQUEST$ to the CHs and records the ones it gets a $JOIN\_RESPONSE$ from by setting corresponding $CONNECTION\_CH$ to true. Let $\theta_{CH,i}$ be the FL model parameter of CH at $i$th FL iteration. In the $JOIN\_RESPONSE$ message, $\theta_{CH,i}$ is also included to make it available for every vehicle to find the cosine similarity between the FL model parameters. The vehicle should choose which CH is the best to join by considering both the speed difference between the vehicles and the similarity of the FL model parameters by using the metric given by
\begin{multline*} \label{eq:1}
Avg\_CoSim_{CH} = (\alpha (|Speed_{CH}-Speed_{k}|)\\
+ (1 - \alpha)(1 - Cosine\_Similarity (\theta_{CH,i},\theta_{k,i})))\tag{1}
\end{multline*} 
where $Speed_{CH}$ and $Speed_{k}$ are the speeds of $CH$ and $k_{th}$ vehicle, respectively; $k$ is the index of the current vehicle; $\alpha$ is the weighting factor between 0 and 1; $Cosine\_Similarity(\theta_{CH,i},\theta_{k,i})$ is the cosine similarity between the parameters of the FL model given by
\begin{equation} \label{eq:2}
Cosine\_Similarity\left( {\theta_{CH,i},\theta_{k,i}} \right) = \frac{{\left\langle {{\theta _{CH,i}},{\theta _{k,i}}} \right\rangle }}{{\left\| {{\theta _{CH,i}}} \right\|\left\| {{\theta _{k,i}}} \right\|}}\tag{2}
\end{equation}
where $\theta_{k,i}$ is the FL model parameter of $k$th vehicle at $i$th FL iteration, and the dot product is a sum of the element-wise products of two vectors (Lines 3-9). Since the cosine distance metric is not affected by scaling effects, it measures the closeness of the direction of two vectors~\cite {9716076}.
Subsequently, the CH with the minimum value of the $Avg\_CoSim_{CH}$ is selected and sent the $CONNECT\_REQUEST$ (Lines 10-12). Receiving a $CONNECT\_RESPONSE$ shows that the $CONNECT\_REQUEST$ was received successfully. As a result, the vehicle switches from state $SE$ to state $CM$ (Lines 13-15). If no $CONNECT\_RESPONSE$ is received, the connection to that CH is set to false  (Lines 16-17). Then, the next CH from the set of the remaining CHs is checked by sending the $CONNECT\_REQUEST$ (Lines 10-12). This continues until the vehicle can connect to one of the CHs.

If there is no CH in one hop or the vehicle cannot connect to any of the CHs in one hop, the vehicle tries to connect to one of the CMs within one hop with the maximum number of hops to the CH less than $MAX\_HOP-1$ (Lines 20-34). Let $\theta_{CM,i}$ be the FL model parameter of CM at $i$th FL iteration. First, the vehicle initializes the $CONNECTION_{CM}$ to false and then sends the $JOIN\_REQUEST$ to the CMs and keeps the ones the response came from by setting corresponding $CONNECTION\_CM$ to true. $JOIN\_RESPONSE$ includes $\theta_{CM,i}$. The selection of CM to connect is based on computing $Avg\_CoSim_{CM}$, same as (1) by replacing CH with CM (Lines 18-26). The CM with the lowest $Avg\_CoSim_{CM}$ value is selected and sent the $CONNECT\_REQUEST$ packet. If the vehicle gets the $CONNECT\_RESPONSE$, it connects to the CM, and the vehicle's status is set to $CM$. If the connection to a CM is not possible, the $CONNECTION_{CM}$ is set to false (Lines 27-34). Then, the next CM from the set of the remaining CMs is checked by sending the $CONNECT\_REQUEST$ (Lines 27-29). This continues until the vehicle can connect to one of the CMs.

If the vehicle fails to connect to any surrounding CH or CM, and there is at least one adjacent vehicle in state $SE$ in the $VIB$, the vehicle calculates the $Avg\_CoSim_{k}$ which is the weighted sum of the average relative speed with respect to all the neighboring nodes and cosine similarity between the previous and the current model parameter as
\begin{multline*} \label{eq:1}
Avg\_CoSim_{k} = (\alpha (Average\_Speed_{k})\\
+ (1 - \alpha)(1 - Cosine\_Similarity (\theta_{k,i},\theta_{k,i-1})))\tag{3}
\end{multline*}
and $Average\_Speed_{k}$ is calculated by
\begin{equation} \label{eq:3}
Average\_Speed_{k} = \frac{\sum_{j=1}^{N(k)}|S_{k} - 
S_{j}|}{N(k)}\tag{4},
\end{equation}
where $N(k)$ is the number of same-direction neighbors for vehicle $k$ within $MAX\_HOP$ hops, $j$ is the $j$th same direction neighbor of vehicle $k$ and $S_k$ is the speed of the vehicle $k$.
The reason for using the average relative speed in the selection of the CH is to achieve cluster stability such that the changes in the CHs are minimized. On the other hand, the reason for the selection of the cosine similarity between the previous and current model parameters at the same vehicle is that the similarity of the FL parameters at the same node shows the vehicle's capability for training and how reliable the training procedure is. Moreover, this eliminates the overhead of collecting the FL parameters from the neighboring vehicles in the $SE$ state. Since the cosine similarity of the model parameters and average relative speed are included in $HELLO\_PACKET$, $Avg\_CoSim_{SE}$ can be calculated for each SE in VIB. The vehicle switches to state $CH$ if it is the one with the minimum value of $Avg\_CoSim_{SE}$ (Lines 35-39).

\begin{algorithm}
Initialize $Timer\_CH$ to 0\;
\For { all $CH \in VIB$ in one hop}
    {
        $Avg\_CoSim_{CH} = (\alpha (Average\_Speed_{CH}) + (1 - \alpha)(1 - Cosine\_Similarity (\theta_{CH,i},\theta_{CH,i-1})))$\;
        Add $Avg\_CoSim_{CH}$ to $VIB$\;
    }
    $CH_{Connect} =$ $\{$ $CH$ $\in$ $VIB$ $\mid$ in one hop\ $\}$\;
    \While {$CH_{Connect} \neq \emptyset$}{
        $CH_{k}= \underset{CH \in CH_{Connect}}{\operatorname{argmin}} (Avg\_CoSim_{CH})$\;
        \If {$Avg\_CoSim_{CH_k} < Avg\_CoSim_{k}$}
        {
            Send $CONNECT\_REQUEST$ to $CH_k$\;
            \If {$CONNECT\_RESPONSE$ received}
            {
	       $STATE = CM$\;
              Exit\;
            }
            \Else{
            $CH_{Connect} = CH_{Connect} - CH_k$\;
            }
        }
        \Else{
        Exit\;
        }
    
    }
    \While {$STATE = CH$}{
        \If {$JOIN\_REQUEST$ OR $g_{k,i}$ not received from any $SE$ or $CM$ $\in$ $VIB$}{
            $Timer\_CH$++ at every $time$\;
            \If {$Timer\_CH> TIMER\_CH\_MAX$}
            {
            $STATE = SE$\;
            Exit\;
            }
        }
        \Else{
        $Timer\_CH = 0$\;
        }
    }
      
\caption{$CH$ State in Multi-hop FL-based Clustering in
VANETs}
\end{algorithm}

To minimize the number of CHs in the network, the CH may connect to another CH with a better $Avg\_CoSim_{CH}$ metric over time, as shown in Algorithm 3. If a vehicle is a CH, it computes $Avg\_CoSim_{CH}$ for all the CHs in one hop (Lines 1-4). To identify the most stable CH among the neighboring CHs, the CH needs to compute the $Avg\_CoSim_{CH}$ by utilizing the current and the most recent $\theta_{CH}$ values. Utilizing the average relative speed for CH selection aims to establish a state of cluster stability, effectively minimizing alterations in the CH composition. Conversely, the adoption of cosine similarity, applied to the comparison between previous and current model parameters within a single vehicle, serves to measure the coherence of FL parameters at a specific node. This choice of metric reflects the vehicle's training proficiency and the reliability of the training process itself. If the minimum $Avg\_CoSim_{CH}$ among the neighboring CHs within one hop is less than that of the current CH, the vehicle sends a $CONNECT\_REQUEST$. If the vehicle receives $CONNECT\_RESPONSE$ successfully, its state changes to $CM$. Otherwise, that CH is removed from the available CH list, and the $Avg\_Cosim_{CH}$ of the remaining CHs are checked (Lines 5-14). The procedure continues until a better CH has been found. If no such CH exists, the vehicle stays in the $CH$ state (Lines 15-16). Let $g_{k,i}$ be the gradient of $k$th vehicle. In the absence of receiving any $JOIN\_REQUEST$ from vehicles in $SE$ states, or if no $g_{k,i}$ is received from the CMs, indicating a loss of connection to all CMs, an increment will occur in the $Timer\_CH$ at every unit time. If this value exceeds the threshold $TIMER\_CH\_MAX$, the vehicle state transitions to $SE$ (Lines 17-23).

\begin{algorithm}
Initialize $Timer\_CM$ to 0\;
    \While {$STATE = CM$}{
        \If {$\theta_{CH,i}$ not received}{
            $Timer\_CM$++ at every $time$\;
            \If {$Timer\_CM> TIMER\_CM\_MAX$}
            {
            $STATE = SE$\;
            Exit\;
            }
        }
        \Else{
        $Timer\_CM = 0$
        }
    }
\caption{$CM$ State in Multi-hop FL-based Clustering in
VANETs}
\end{algorithm}

A vehicle stays in state $CM$ as long as $\theta_{CH,i}$ is received, as shown in Algorithm 4. If $\theta_{CH,i}$ is received, it means that the connection to CM or CH is still valid. Otherwise, it means that the connection to CH or CM may have been lost, so $Timer\_CM$ is updated at every time. If $Timer\_CM$ exceeds the predefined $TIMER\_CM\_MAX$, the connection to the CH is considered lost, and the vehicle switches to state $SE$ (Lines 2-7).

\section{Hierarchical FL over Clustered VANET algorithm}
We propose the usage of the HFL algorithm over the multi-hop clustered architecture, described in Section III, in contrast to the conventional FL (CFL) due to the advantages of distributed learning and computing in multiple clusters. The proposed HFL is capable of dealing with the divergence possibility in non-IID datasets and the mobility of vehicles by using both the mobility measure and the similarity between the latest learning model parameters in FL in the clustering metric. Moreover, it introduces a mechanism to adapt FL to CH changes during the execution of the FL.

In conventional FL, every CM constructs a local gradient vector during each communication round using its private dataset and the global model from the previous communication round. The gradients are then transmitted to the EPC through the respective CHs. At EPC, a global model is built utilizing both the received gradients and the previous communication round global model. The EPC then communicates the global model to all CMs through their connected CHs. This round of updating the global model continues until the global convergence.

In HFL, the local gradient vectors calculated at all the CMs connected to the CHs are aggregated by averaging at the CHs; CHs provide the aggregated model parameters to the EPC. The EPC then generates the global model by averaging the model parameters delivered by CHs and sends it to the CHs to update the model parameters based on the aggregated global model. The updated model parameters are finally sent to the CMs to continue the learning procedure based on the new parameters they received.

The HFL algorithm is provided in Algorithms 5-9 for multihop-clustered VANET architecture in each state. If the vehicle is in state $IN$, it should decide to collect data or start the learning procedure as shown in Algorithm 5. Initially, the vehicle sets both $i$ and $t\_DATA$ to zero, representing the communication round and the time spent on data collection, respectively. Additionally, $\theta_{k,0}$ and $g_{k,0}$ are initialized with random numbers to initiate the FL learning process (Lines 1-3). The values of $t\_DATA$ increment at every time unit (Line 5). For the initial duration of $t\_DATA\_MAX$, the vehicle collects the raw data from the environment (Lines 6-7). After $t\_DATA\_MAX$ duration, the vehicle updates $g_{k,i+1}$ based on raw local datasets by computing Stochastic Gradient Descent (SGD) as given by
\begin{equation}\label{eq:4}
g_{k,i+1} = \nabla_{\theta}F_k(\theta_{k,i},\xi_{k}),\tag{5}
\end{equation}
where $\nabla_{\theta}$ represents the gradient calculated with respect to the variable $\theta$, $F_{k}$ is the loss function of the $k_{th}$ vehicle, and the $\xi_{k}$ is the local dataset for $k_{th}$ vehicle (Lines 9-10). Subsequently, the variable $i$ increments with each FL iteration, leading to an update of $\theta_{k,i}$ based on the model parameters according to the following (Lines 11-13)
\begin{equation}\label{eq:4}
\theta_{k,i} = \theta_{k,i-1} - \eta g_{k,i},\tag{6}
\end{equation}

When the vehicle is in the state $SE$, its responsibility entails computing the SGD and subsequently updating the value of $\theta_{k,i}$ using the most recent $\theta_{k}$, as shown in Algorithm 6. This update is instrumental in the calculation of $Avg\_CoSim_{SE}$ in Algorithm 2.

\begin{algorithm}

Initialize $i$ to 0\;
Initialize $t\_DATA$ to 0\;
Initialize $g_{k,0}$ and $\theta_{k,0}$ to random numbers\;
\While{$STATE = IN$}{
    $t\_DATA$++ at every $time$\;
    \If{$t\_DATA$ $<$ $t\_DATA\_MAX$}{
        Collect raw local datasets}
    \Else{
        Update $g_{k}$ based on raw local datasets by\\
        $g_{k,i+1}$ = $\nabla_{\theta}F_k(\theta_{k,i},\xi_{k})$\;
        $i$++\;
        $t\_DATA = 0$\;
        $\theta_{k,i} = \theta_{k,i-1} - \eta g_{k,i}$\;
    }
}

\caption{$IN$ State for HFL in Multi-hop-Clustered VANETs}
\end{algorithm}

\begin{algorithm}
\For {every $i$}{
    Compute Stochastic Gradient Descent (SGD):\\
    $g_{k,i}$ = $\nabla_{\theta}F_{k}(\theta_{k,i-1},\xi_{k})$\;
    $\theta_{k,i} = \theta_{k,i-1} - \eta g_{k,i}$\;
    }
   
\caption{$SE$ State for HFL in Multi-hop-Clustered VANETs}
\end{algorithm}

If a vehicle is in state $CH$, it is responsible for collecting the gradients from its own CMs, averaging them to obtain model parameters, and forwarding the model parameters to EPC or CMs, as shown in Algorithm 7. First, the CH initializes the time duration for collecting the gradients, $t\_COLLECT\_CH$, to zero. If $t\_COLLECT\_CH$ is less than the predefined $t\_COLLECT\_MAX$, the CH continues collecting all of the $g_{CM,i}$ from the CMs in the set of $\mathcal{CM}$ (Lines 1-5). Otherwise, if the $t\_COLLECT\_CH$ exceeds the $t\_COLLECT\_MAX$, for all the CMs in its set, the model parameter is updated as
\begin{equation}\label{eq:5}
\theta_{CM,i} = \theta_{CH,i-1} - \eta g_{CM,i},\tag{7}
\end{equation}
where $\theta_{CM,i}$ and $\theta_{CH,i-1}$ are the model parameter of CM and the previous model parameter of the CH, respectively, and $\eta$ is the step size in SGD (Lines 6-7). Then, the CH takes the average of the CM model parameters based on the FedAVG algorithm in the set of $\mathcal{CM}$, as given by
\begin{equation}\label{eq:6}
\theta_{CH,i} = \sum_{CM \in \mathcal{CM}} \frac{ n_{CM}}{{n}} \theta_{CM,i}\tag{8}
\end{equation}
where $\theta_{CH,i}$ is the CH model parameter; ${n_{CM}}$ is the number of dataset samples available to CM, and $n$ is the total number of data samples for all CM$\in$$\mathcal{CM}$. The CH then forwards $\theta_{CH,i}$ to EPC (Lines 8-9). The CH waits until the model parameter of the EPC, $\theta_{EPC,i}$, is received, which then updates the $\theta_{CH,i}$ (Lines 10-11). The $\theta_{CH,i}$ is transmitted to all of the CMs in the $\mathcal{CM}$ set for the following iteration of the FL procedure (Line 12).

\begin{algorithm}

\For {every $i$ }{
    Initialize $t\_COLLECT\_CH$ to 0\;
    $t\_COLLECT\_CH$++ at every $time$\;
    \While {$t\_COLLECT\_CH<t\_COLLECT\_MAX$}{
        Collect $g_{CM,i}$ from all CMs $\in \mathcal{CM}$\;
    }
    \For {all $CM \in \mathcal{CM}$}
        {
        $\theta_{CM,i} = \theta_{CH,i-1} - \eta g_{k,i}$\;
        $\theta_{CH,i} = $ $\sum_{CM \in \mathcal{CM}}$ $\frac{ n_{CM}}{{n}} \theta_{CM,i}$\;
        }
        Forwards $\theta_{CH,i}$ to its EPC\;
        Wait until {$\theta_{EPC,i}$ Received}\\
        {
        $\theta_{CH,i} = \theta_{EPC,i}$\;
        }
        Send $\theta_{CH,i}$ to all CMs $\in \mathcal{CM}$\;
    }
\caption{$CH$ State for HFL in Multi-hop-Clustered VANETs}
\end{algorithm}

If a vehicle is in state $CM$, it computes the gradients based on the model parameter sent by its corresponding CH, as shown in Algorithm 8. When $\theta_{CH,i}$ is received, the CM computes the SGD based on that and sends the latest $g_{CM,i}$ to the CH (Lines 1-4).

\begin{algorithm}
\For {every $i$}{
    Compute Stochastic Gradient Descent (SGD):\\
    $g_{CM,i}$ = $\nabla_{\theta}F_{k}(\theta_{CH,i},\xi_{k})$\;
    Send $g_{CM,i}$ to CH\;
}
    
\caption{$CM$ State for HFL in Multi-hop-Clustered VANETs}
\end{algorithm}

Finally, at the EPC, a global model is constructed by combining the model parameters collected from the CHs with the global model derived in the previous communication round by using the FedAVG algorithm, as shown in Algorithm 9. If the FL convergence speed exceeds a predetermined threshold denoted by $\epsilon$, the EPC resets the timer for collecting model parameters, denoted by $t\_COLLECT\_EPC$, to zero. If the elapsed time, $t\_COLLECT\_EPC$, is less than the predefined threshold, $t\_COLLECT\_MAX$, the EPC proceeds to collect all $\theta_{CH,i}$ values from the CHs within the set $\mathcal{CH}$ (Lines 1-6). If $t\_COLLECT\_EPC$ exceeds $t\_COLLECT\_MAX$ threshold, the global model parameter is computed using the model parameters obtained from the CHs in the set by
\begin{equation}\label{eq:7}
\theta_{EPC,i} = \theta_{EPC,i-1}  - \eta\sum_{CH \in \mathcal{CH}}\frac{ n_{CH}}{{n}} \theta_{CH,i},\tag{9}
\end{equation}
where $\theta_{EPC,i}$ is the EPC global model parameter; ${n_{CH}}$ is the number of dataset samples available to CH, and $n$ is the total number of samples in all CHs in the set $\mathcal{CH}$ (Lines 7-8). Subsequently, the most recent global model parameter is transmitted to all the CHs (Line 9), and following this, it is conveyed to all CMs through their corresponding CHs. The learning continues until the EPC global model's convergence speed falls below a particular threshold.

\begin{algorithm}
\While{FL convergence speed $\geq$ $\epsilon$}{
    \For {every $i$ }{
        Initialize $t\_COLLECT\_EPC$ to 0\;
        $t\_COLLECT\_EPC$++ at every $time$\;
        \While {$t\_COLLECT\_EPC<t\_COLLECT\_MAX$}{
            Collect $\theta_{CH,i}$ from all CHs $\in \mathcal{CH}$\;
        }
        \For {all $CH \in \mathcal{CH}$}{
            $\theta_{EPC,i}$ = $\theta_{EPC,i-1}  - \eta$ $\sum_{CH \in \mathcal{CH}}$ $\frac{ n_{CH}}{{n}} \theta_{CH,i}$\;
            Send $\theta_{EPC,i}$ to all CHs $\in \mathcal{CH}$\;
        }
    }
}
\caption{EPC for HFL in Multi-hop-Clustered VANETs}
\end{algorithm}

\begin{table}[h!]
\caption{Description of the variables}
\centering
\begin{tabular}{|l|l|lll}
\cline{1-2}
\textbf{Notation} & \textbf{Description}  \\ \cline{1-2}
i                 & FL iteration \\ \cline{1-2}
$\epsilon$        & FL convergence threshold \\ \cline{1-2}
$\mathcal{CM}$    & Set of all CMs under a CH \\ \cline{1-2}
$\mathcal{CH}$    & Set of all CHs under a EPC \\ \cline{1-2}
$F_k$             & Loss function of vehicle $k$ \\ \cline{1-2}
$\theta$          & Model parameters \\ \cline{1-2}
$\xi_{k}$         & Local dataset of vehicle $k$ \\ \cline{1-2}
CH                & Cluster Head \\ \cline{1-2}
CM                & Cluster Member \\ \cline{1-2}
$n$               & Number of all data points \\ \cline{1-2}
$n_{CM}$          & Number of data points available to $CM$ \\ \cline{1-2}
$n_{CH}$          & Number of data points available to $CH$ \\ \cline{1-2}
$g_{k,i}$         & Gradient of node $k$ at $i$th iteration \\ \cline{1-2}
$\eta$            & Step size in SGD \\ \cline{1-2}
\end{tabular}
\end{table}

\section{Performance Evaluation}

The aim of the simulations is to analyze the performance of the proposed HFL over clustered VANET algorithm for different network size and transmission range values and compare it to that of three existing approaches: Traditional FL without the usage of the clustering (No Clustering)~\cite{9360666} HFL over previously proposed Multihop-Cluster-Based IEEE 802.11p and LTE Hybrid Architecture (VMaSC)~\cite{7081788}, and HFL over previously proposed clustering algorithm for FL based on the similarity of their local updates (CVFL)~\cite{9716076}. In traditional FL, clustering is not incorporated, and all vehicles are treated as clients, while an EPC is considered as the server and functions as the central server. In VMaSC, the selection of a CH for cluster formation relies solely on the average relative speed metric. On the other hand, in CVFL, the CH selection and cluster creation are determined exclusively by evaluating cosine similarity.

\subsection{Simulation Setup}

\begin{figure*}[ht]
\setlength\belowcaptionskip{0pt}
\centering
\centerline{\includegraphics[width=1.2\textwidth]{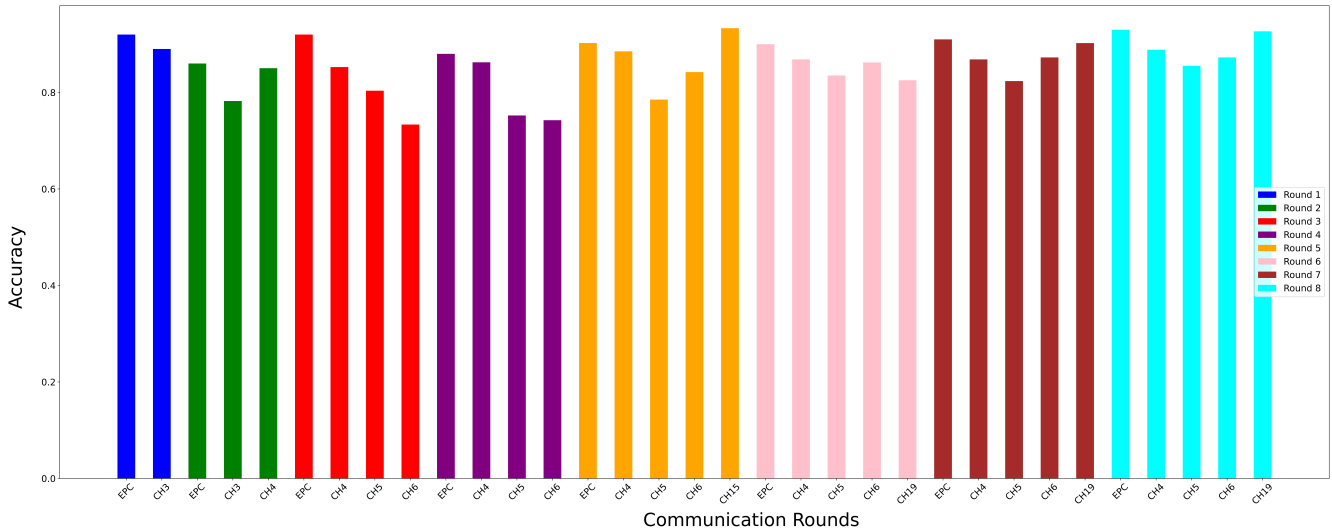}}
\centering
\caption{Accuracy progression of CHs with $\alpha= 0.5$ and $MAX\_HOP = 1$}
\end{figure*}

\begin{figure}[h]
     \centering
     \begin{subfigure}{0.45\textwidth}
         \centering
         \includegraphics[width=\textwidth]{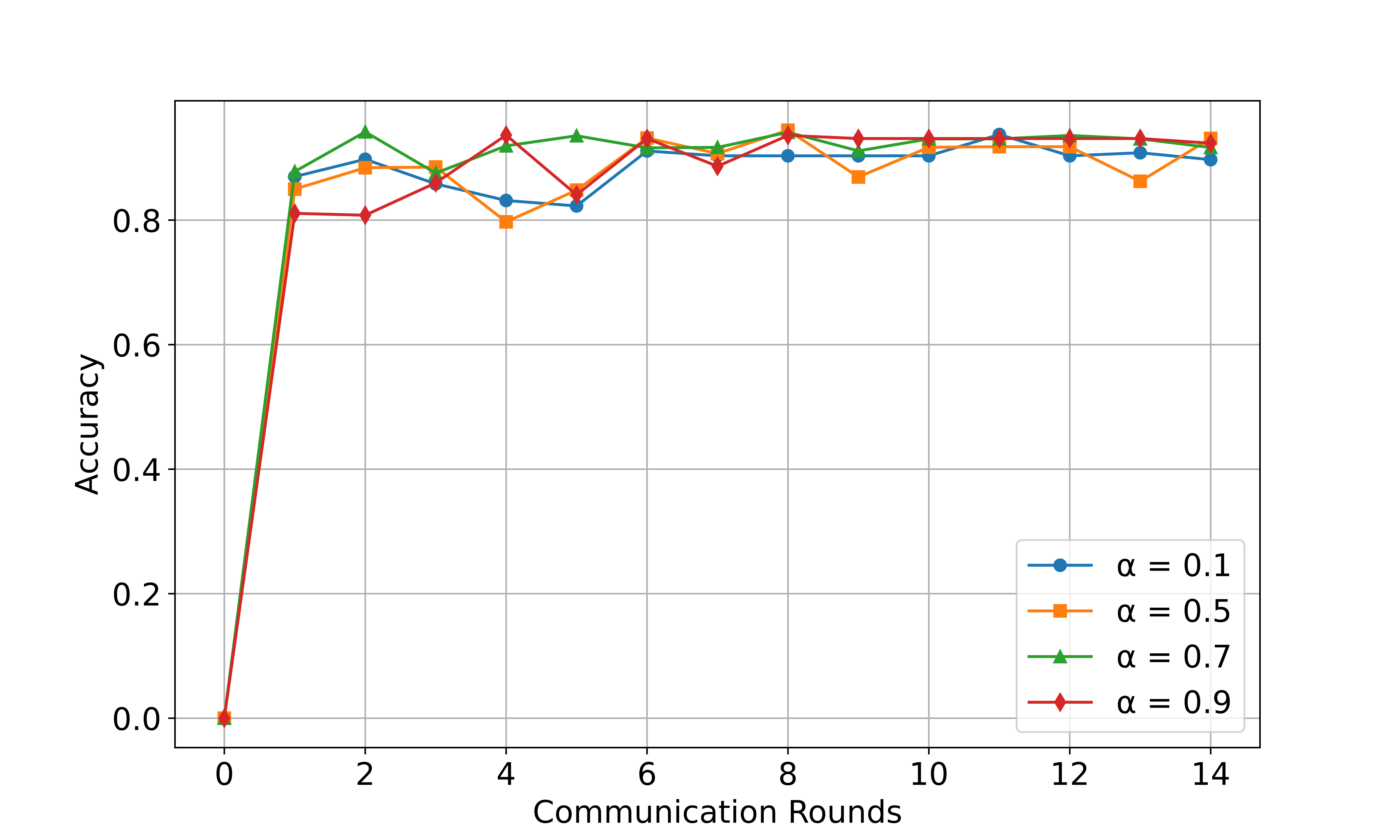}
         \caption{10 Vehicles, Tx Range= 100 Meters}
         \label{fig: 10-100}
     \end{subfigure}
     \vfill
     \begin{subfigure}{0.45\textwidth}
         \centering
         \includegraphics[width=\textwidth]{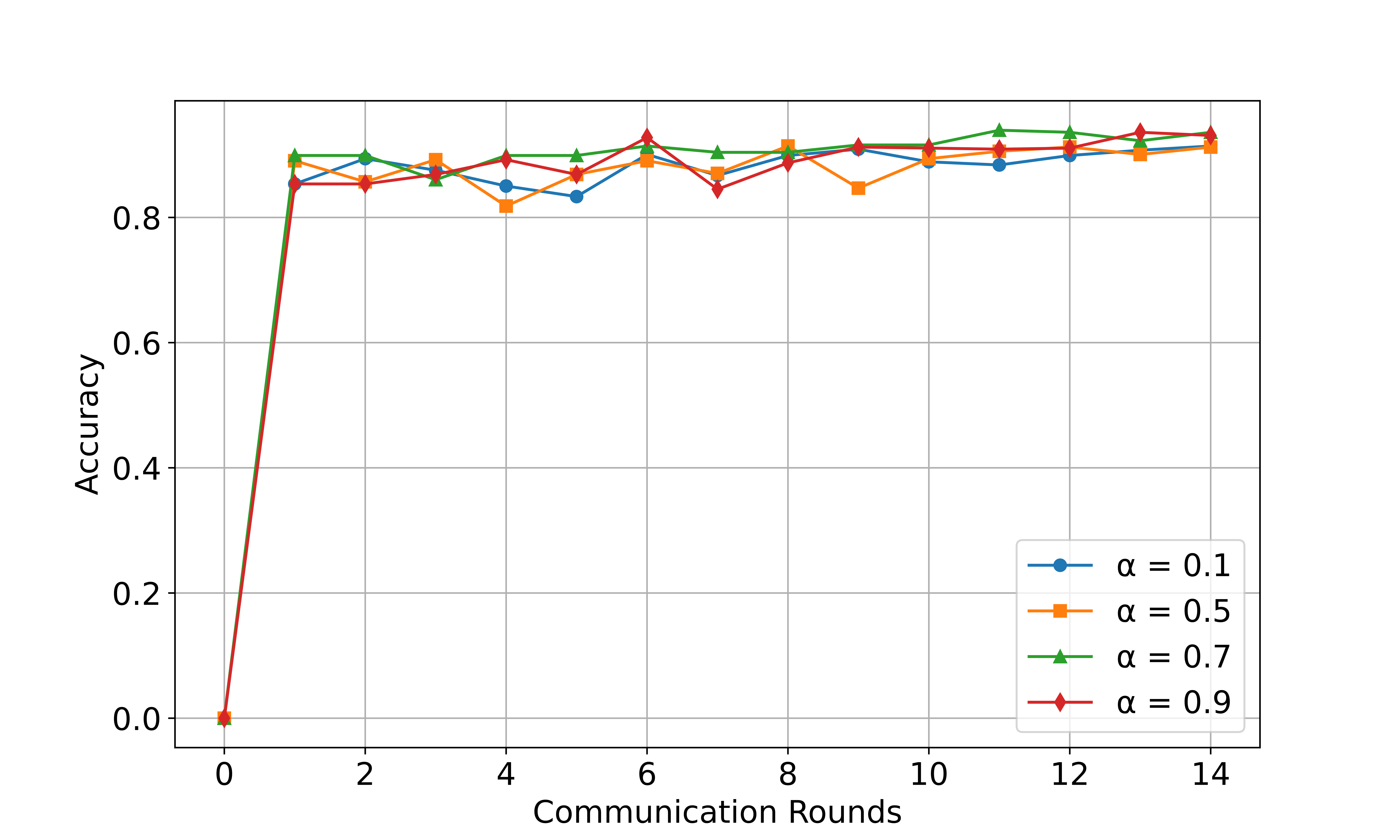}
         \caption{20 Vehicles, Tx Range= 100 Meters}
         \label{fig: 20-100}
     \end{subfigure}
     \vfill
     \begin{subfigure}{0.45\textwidth}
         \centering
         \includegraphics[width=\textwidth]{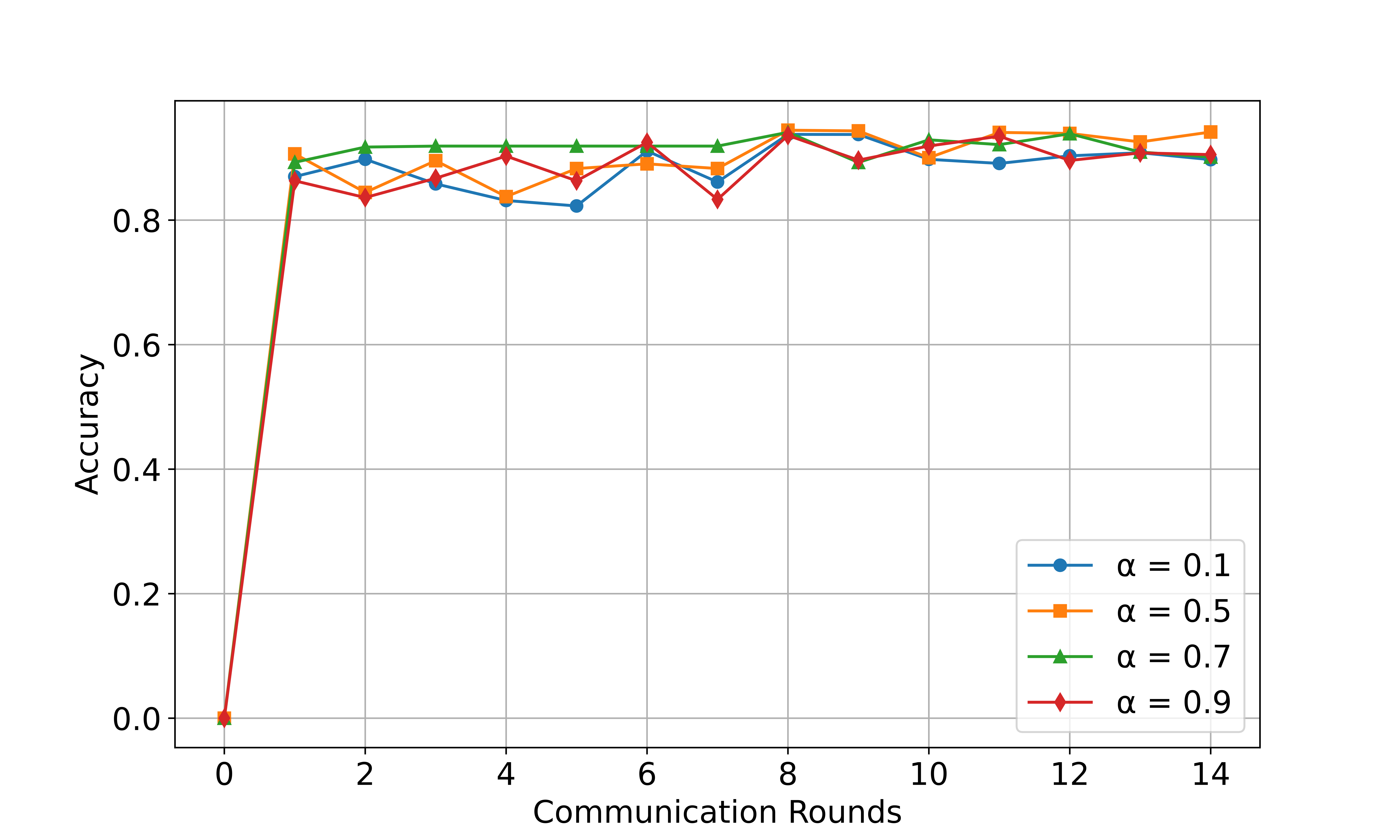}
         \caption{20 Vehicles, Tx Range= 500 Meters}
         \label{fig: 20-500}
     \end{subfigure}
        \caption{The accuracy in the EPC for different numbers of nodes and transmission ranges using various $\alpha$ values with $MAX\_HOP = 1$.}
        \label{fig: EPC Accuracy for MAX_HOP=1}
\end{figure}

\begin{figure}[h]
     \centering
     \begin{subfigure}{0.45\textwidth}
         \centering
         \includegraphics[width=\textwidth]{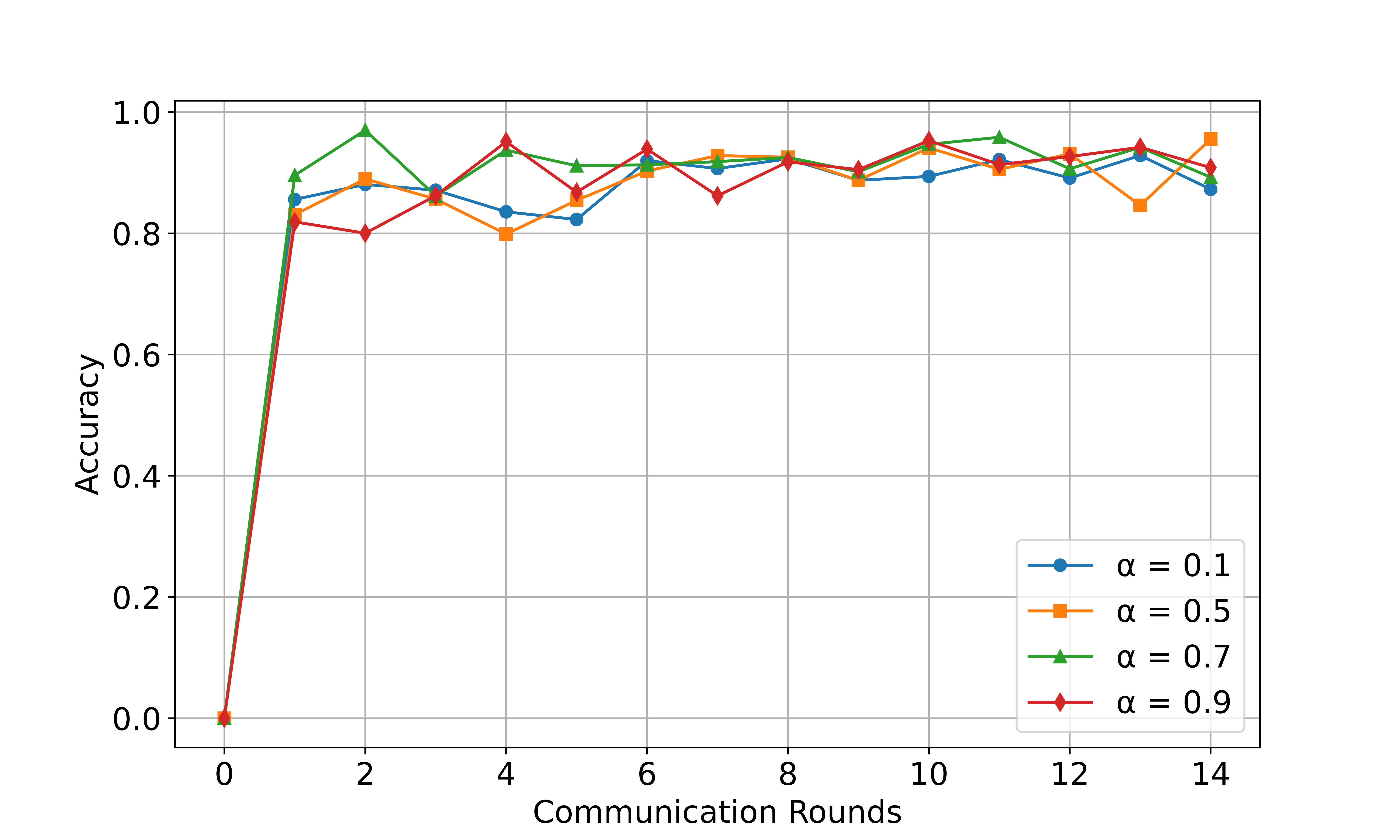}
         \caption{10 Vehicles, Tx Range= 100 Meters}
         \label{fig: 10-100}
     \end{subfigure}
     \vfill
     \begin{subfigure}{0.45\textwidth}
         \centering
         \includegraphics[width=\textwidth]{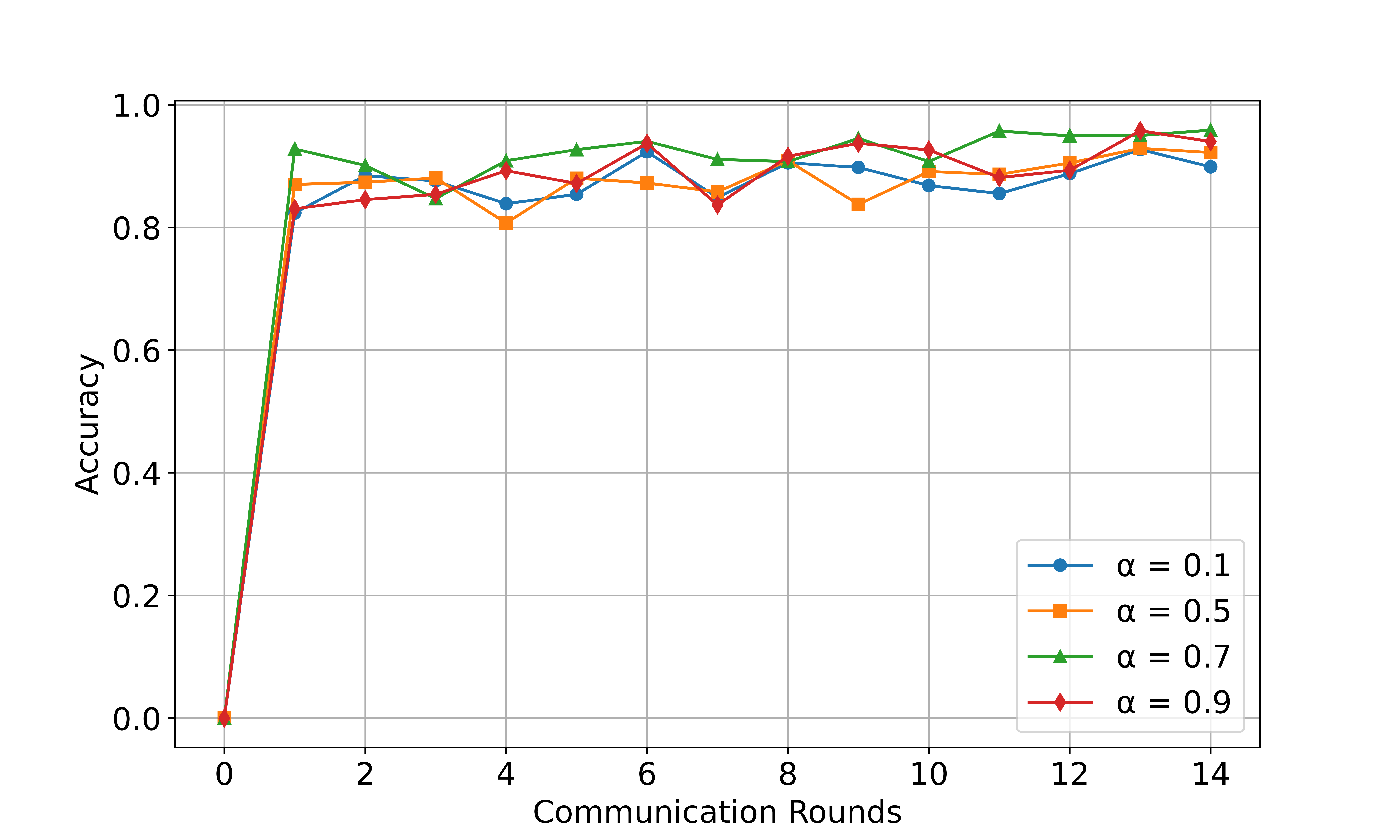}
         \caption{20 Vehicles, Tx Range= 100 Meters}
         \label{fig: 20-100}
     \end{subfigure}
     \vfill
     \begin{subfigure}{0.45\textwidth}
         \centering
         \includegraphics[width=\textwidth]{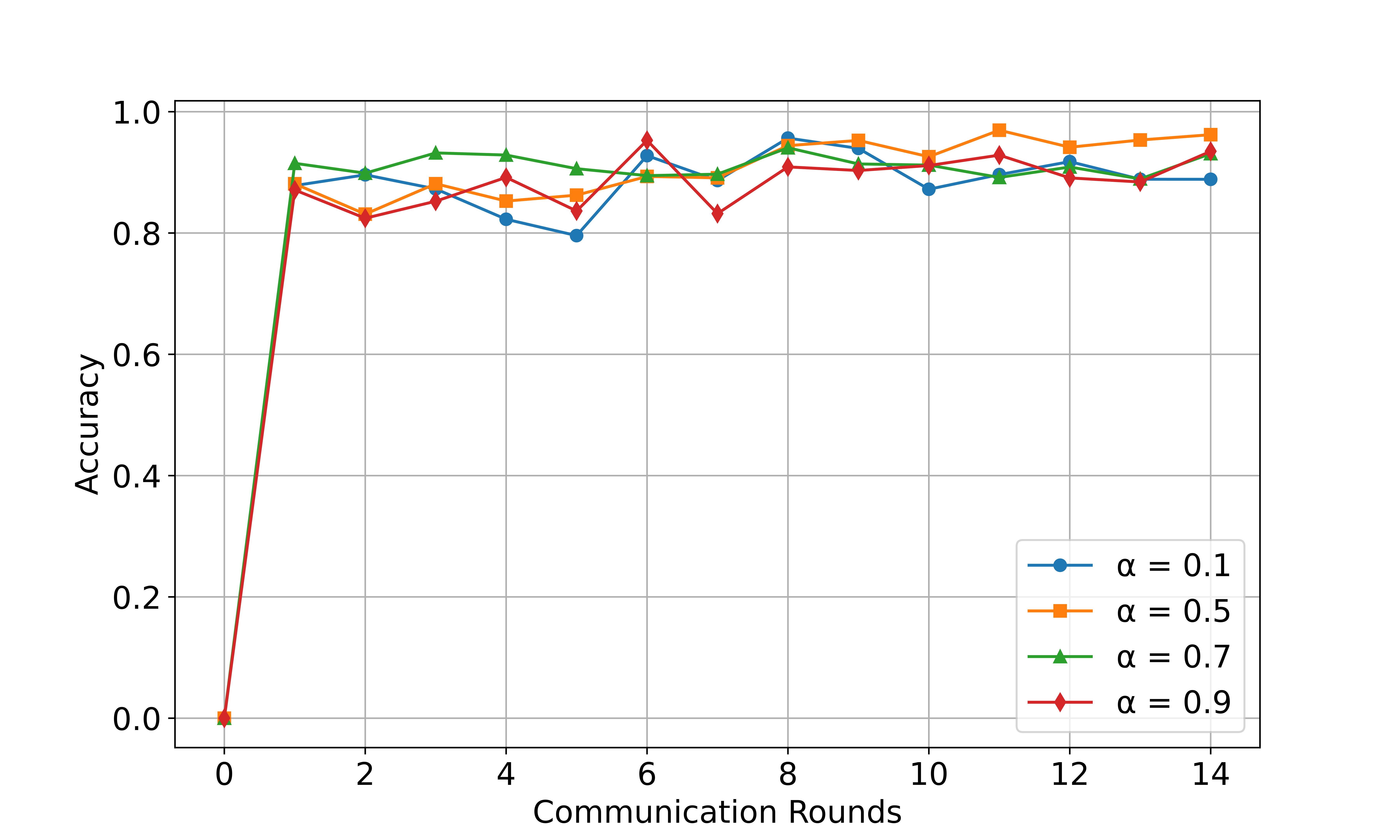}
         \caption{20 Vehicles, Tx Range= 500 Meters}
         \label{fig: 20-500}
     \end{subfigure}
        \caption{The accuracy in the EPC for different numbers of nodes and transmission ranges using various $\alpha$ values with $MAX\_HOP = 2$.}
        \label{fig: EPC Accuracy for MAX_HOP=2}
\end{figure}

\begin{figure}[h]
     \centering
     \begin{subfigure}{0.45\textwidth}
         \centering
         \includegraphics[width=\textwidth]{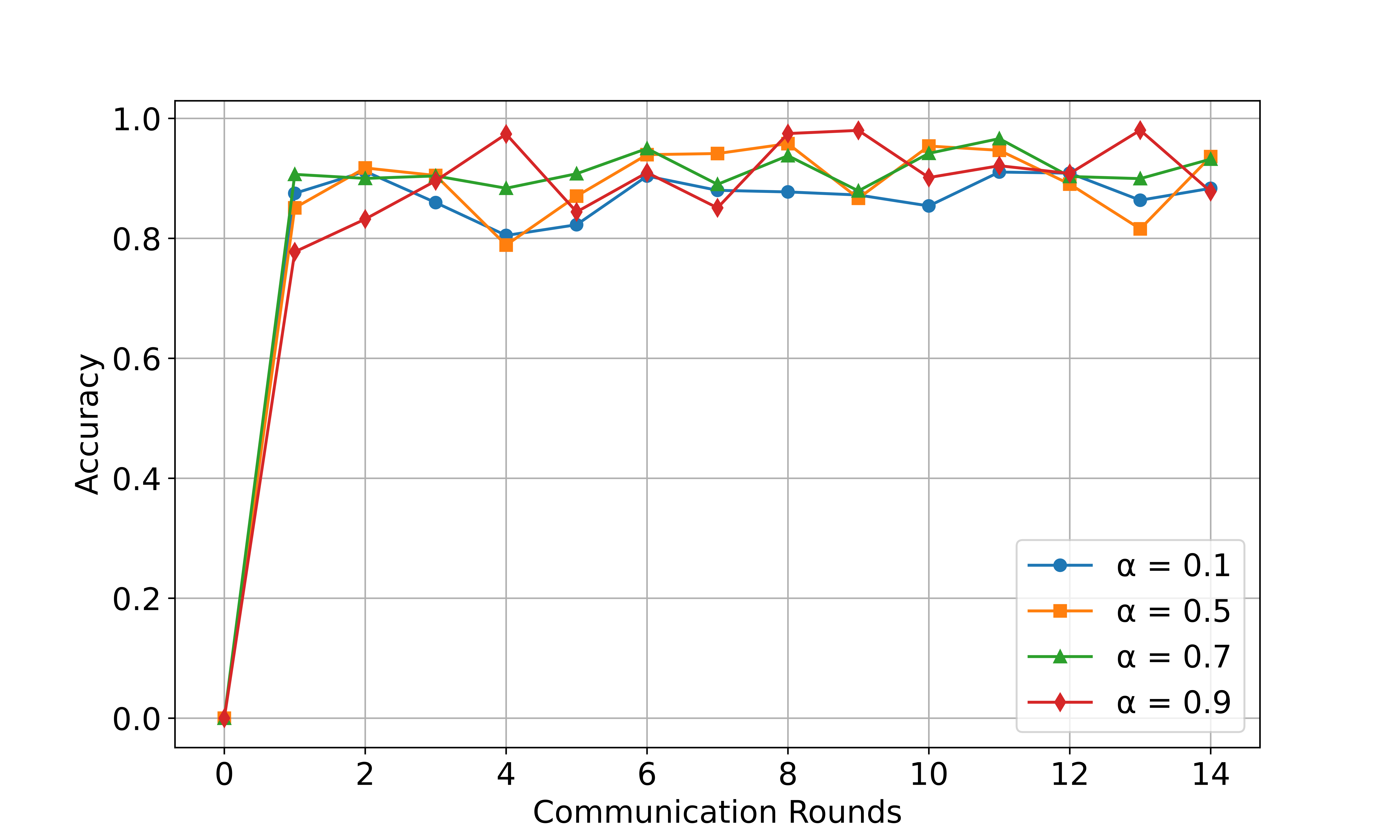}
         \caption{10 Vehicles, Tx Range= 100 Meters}
         \label{fig: 10-100}
     \end{subfigure}
     \vfill
     \begin{subfigure}{0.45\textwidth}
         \centering
         \includegraphics[width=\textwidth]{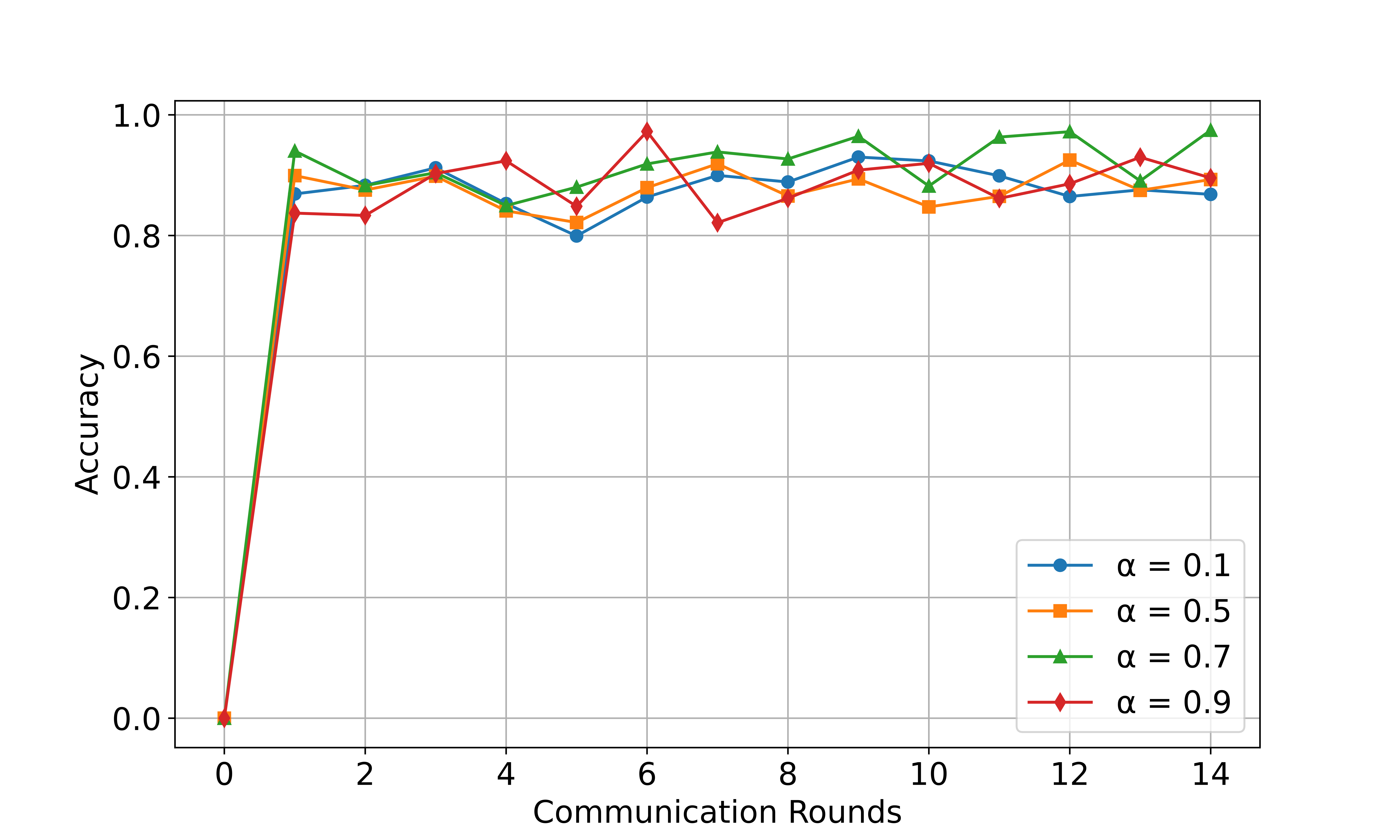}
         \caption{20 Vehicles, Tx Range= 100 Meters}
         \label{fig: 20-100}
     \end{subfigure}
     \vfill
     \begin{subfigure}{0.45\textwidth}
         \centering
         \includegraphics[width=\textwidth]{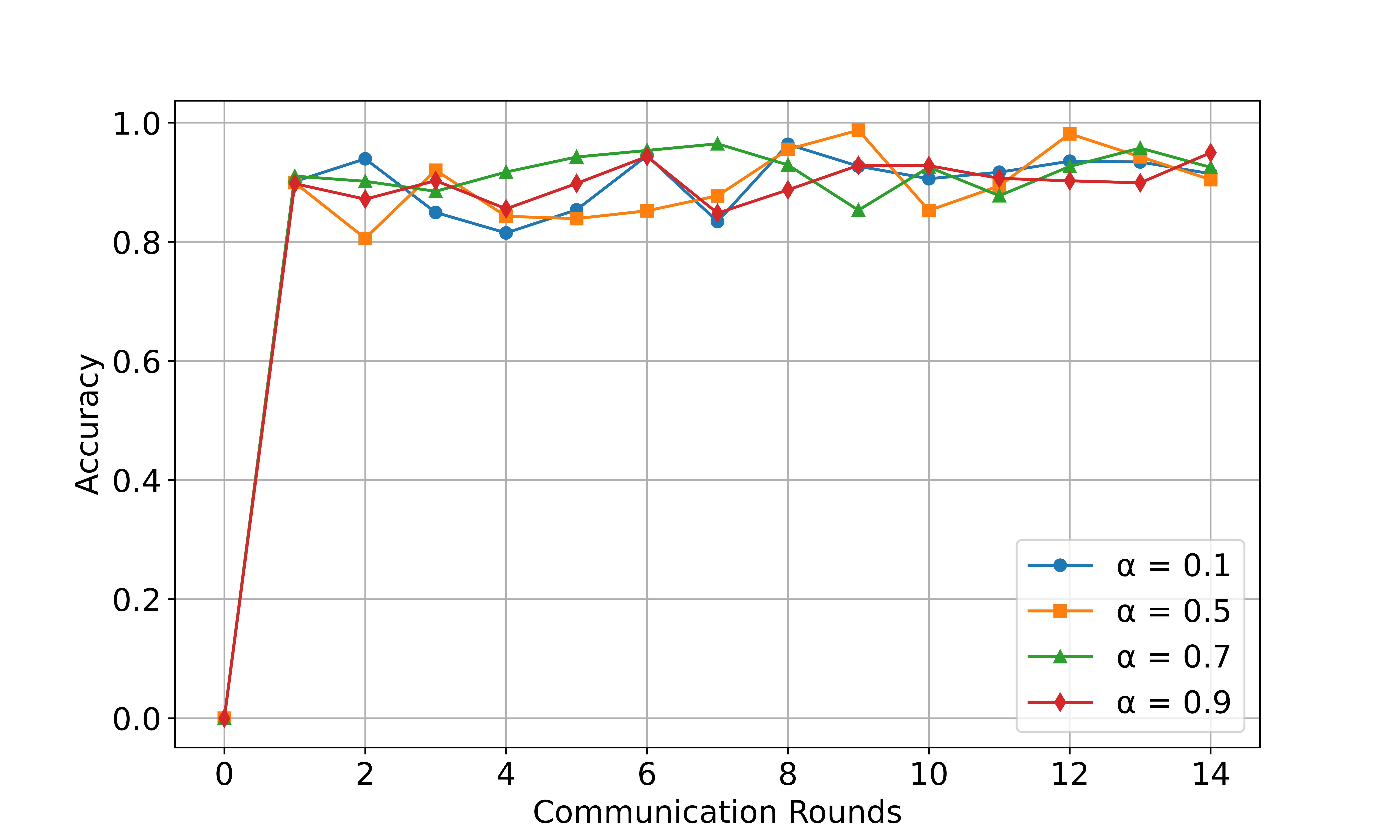}
         \caption{20 Vehicles, Tx Range= 500 Meters}
         \label{fig: 20-500}
     \end{subfigure}
        \caption{The accuracy in the EPC for different numbers of nodes and transmission ranges using various $\alpha$ values with $MAX\_HOP = 3$.}
        \label{fig: EPC Accuracy for MAX_HOP=3}
\end{figure}

The simulations are conducted using Python, incorporating realistic vehicle mobility generated by Simulation of Urban Mobility (SUMO) and wireless data streaming provided by KAFKA. The software for the implementation is available in~\cite{VANETRepository}. SUMO, developed by the German Aerospace Center, and is an open-source traffic simulator that accurately models the behavior of individual drivers while considering factors such as distance to the leading vehicle, travel speed, vehicle dimensions, and acceleration-deceleration profiles to determine vehicle acceleration and overtaking decisions~\cite{SUMO}. The FL framework is implemented using PyTorch~\cite{10.1145/3488659.3493775} or TensorFlow~\cite{TensorFlow}. For packet transmission, KAFKA is utilized to generate and send $HELLO\_PACKET$s and model parameters between the vehicles~\cite{KAFKA}. Apache Kafka is an open-source distributed event streaming platform known for its high-performance data pipelines, streaming analytics, data integration, and support for mission-critical applications.

The road topology comprises a two-lane, two-way road map spanning an area of 1 km$^2$. Vehicles are injected onto the road using a Poisson process. To create a realistic scenario with various types of vehicles, the simulation incorporates vehicles with different maximum speed ranges. The maximum speed of the vehicles varies from 10 m/s to 35 m/s, allowing for a range of speeds on the road. The vehicle density is 10 and 20 vehicles per km$^2$ for different scenarios.

IEEE802.11p is employed for vehicle-to-vehicle (V2V) communication, whereas 5G NR is used for the communication between the CHs and the EPC. The implementation of the Winner+ B1 propagation model proposed in~\cite{9599363} is used for V2V communications, and the Friis propagation model is employed in V2I communication scenarios due to its ability to estimate signal attenuation and propagation characteristics between vehicles and 5G NR base stations~\cite{3gpptr38901}. The transmission range for each vehicle is assumed to be 100 or 500 meters for different scenarios, while the EPC is presumed to have coverage over the entire area, which is 1 km$^2$.

During the FL training process, the vehicles utilize the well-known benchmark image classification datasets, namely MNIST~\cite{726791}. MNIST comprises 28x28 grayscale images of handwritten digits with random classes assigned to ensure non-IID data distribution. To introduce diversity, the dataset is randomly partitioned and made available to the vehicles. The dataset consists of 60,000 training examples and 10,000 test examples. Vehicles engage in cooperative training of a multi-layer perceptron (MLP) model consisting of two hidden layers, each containing 64 neurons. Additionally, a convolutional neural network (CNN) model is employed, comprising two 5 × 5 convolutional layers (10 channels in the first and 20 in the second), each followed by 2 × 2 max pooling. The CNN model also integrates two fully connected layers with 50 units and ReLU activation, culminating in a softmax output layer~\cite{9716076}.
The performance metrics used to evaluate the FL are accuracy and convergence time. Accuracy is commonly measured using evaluation metrics that assess the model's performance on a separate validation dataset, which measures the proportion of correct predictions made by the learning model when tested with samples from the dataset~\cite{Goodfellow2016}. The convergence time refers to the duration taken for the distributed learning process to achieve a desired level of performance~\cite{mcmahan2017communication}.

\begin{table*}[ht]
\centering
\caption{Convergence time based on different values of $\epsilon$ and $\alpha$ for  $MAX\_HOP = 1$}
\begin{tabular}{|c|c|c|c|c|c|c|c|c|}
\hline
\multirow{5}{*}{\begin{tabular}[c]{@{}c@{}}20 Vehicles\\ Tx Range = 100 Meters\end{tabular}} &                  & \multicolumn{1}{l|}{No Clustering} & $\alpha=0$ & $\alpha=0.1$ & $\alpha=0.5$ & $\alpha=0.7$ & $\alpha=0.9$ & $\alpha=1$ \\ \cline{2-9} 
                                                                                             & $\epsilon=0.01$  & 14                                 & 4          & 3            & 11           & 6            & 3            & 10         \\ \cline{2-9} 
                                                                                             & $\epsilon=0.005$ & 15                                 & 21         & 11           & 24           & 12           & 10           & 17         \\ \cline{2-9} 
                                                                                             & $\epsilon=0.001$ & 17                                 & 23         & 16           & 29           & 15           & 10           & 17         \\ \cline{2-9} 
                                                                                             & Average          & 15.3333                            & 16         & 10           & 21.3333      & 11           & 7.6667       & 14.6667    \\ \hline
\multirow{4}{*}{\begin{tabular}[c]{@{}c@{}}10 Vehicles\\ Tx Range = 100 Meters\end{tabular}} & $\epsilon=0.01$  & 17                                 & 7          & 6            & 15           & 8            & 6            & 13         \\ \cline{2-9} 
                                                                                             & $\epsilon=0.005$ & 18                                 & 25         & 14           & 27           & 15           & 14           & 20         \\ \cline{2-9} 
                                                                                             & $\epsilon=0.001$ & 20                                 & 27         & 19           & 31           & 19           & 14           & 20         \\ \cline{2-9} 
                                                                                             & Average          & 18.3333                            & 19.6667    & 13           & 24.3333      & 14           & 11.3333      & 17.6667    \\ \hline
\multirow{4}{*}{\begin{tabular}[c]{@{}c@{}}20 Vehicles\\ Tx Range = 500 Meters\end{tabular}} & $\epsilon=0.01$  & 15                                 & 5          & 5            & 13           & 5            & 6            & 13         \\ \cline{2-9} 
                                                                                             & $\epsilon=0.005$ & 18                                 & 24         & 12           & 25           & 14           & 12           & 19         \\ \cline{2-9} 
                                                                                             & $\epsilon=0.001$ & 20                                 & 26         & 18           & 27           & 18           & 12           & 19         \\ \cline{2-9} 
                                                                                             & Average          & 17.6667                            & 18.3333    & 11.6667      & 22.3333      & 12.3333      & 10           & 17         \\ \hline
\end{tabular}
\end{table*}

\subsection{Performance Evaluation of Proposed Algorithm}

\begin{figure}[ht]
\setlength\belowcaptionskip{0pt}
\centering
\includegraphics[width=9 cm]{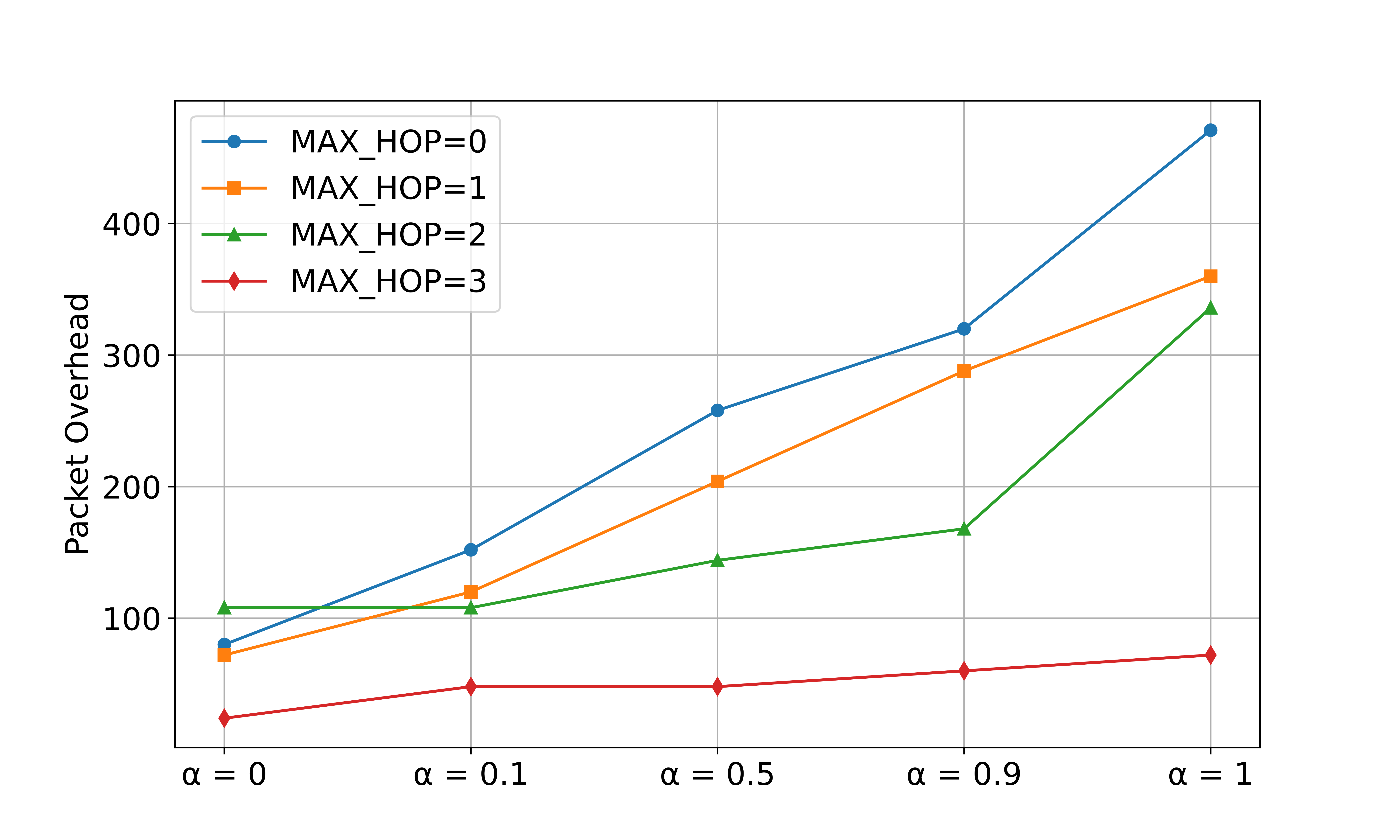}
\caption{The number of packets transmitted to the EPC for different numbers of nodes and $\alpha$s}
\end{figure}

Fig. 2 shows the accuracy progression of CHs over time for $\alpha=0.5$ and a maximum of 1 hop. The simulation involves 20 vehicles with a communication range of 100 meters. The accuracy is represented as a function of the communication round. A communication round is a distinct iteration within the training procedure wherein vehicles interface with a central server or aggregator. Within the figure, the intent is to visualize state transitions facilitated by the clustering algorithm. The variations in accuracy within CHs and EPC occur due to factors such as the introduction of new vehicles, state transitions of existing vehicles, and the merging of multiple CHs. A node's absence in a specific communication round signifies alterations in its state to become a CM or SE or its amalgamation with another CH. Moreover, the variation in accuracy is illustrated across different CHs and the EPC.

Figures 3-5 illustrate the accuracy progression at the EPC over time for different $\alpha$ values and scenarios with different $MAX\_HOP$s. The scenarios encompass the following setups: 1- 10 vehicles with a communication range of up to 100 meters, 2- 20 vehicles with a communication range of up to 100 meters, and 3- 20 vehicles with a communication range of up to 500 meters. The purpose is to determine the optimal $\alpha$ based on the observed accuracy fluctuations, which indicate the susceptibility to convergence loss when new data is introduced. The figures illustrate that $\alpha$ values of 0.9 exhibit lower accuracy fluctuations as the communication rounds progress, converging earlier in the communication process. Furthermore, reducing the number of vehicles at a constant transmission range amplifies fluctuations in the global model, owing to the decreased number of clients in the system. Likewise, maintaining the same number of vehicles while increasing the transmission range intensifies fluctuations due to reduced clustering and fewer CHs; however, this scenario is smoother than the one with ten vehicles due to the higher number of vehicles. Furthermore, an increase in the number of hops for the same scenario causes an increase in the fluctuations. This can be attributed to the fact that a higher number of hops reduces the number of CHs within a cluster, leading to increased fluctuations. This trend is similar to the model with a higher transmission range, resulting in heightened fluctuations. Hence, with an increasing number of vehicles and a simultaneous reduction in transmission range, given an optimal value of $\alpha$, the fluctuation in accuracy diminishes. This effect can be attributed to the higher learning capability and improved adaptability to incoming data within the system. 

Table II presents the convergence time of the proposed algorithm across various values of the desired performance level $\epsilon$ and $\alpha$ for $ MAX\_HOP=1$. The evaluation encompasses distinct scenarios, as mentioned above. The outcomes notably indicate that an $\alpha$ value of 0.9 yields the most favorable convergence time. Additionally, reducing the number of vehicles while maintaining the same transmission range results in an increase in convergence time. This happens because fewer vehicles are available to participate in the FL process as the number of vehicles decreases, causing the learning procedure to take longer compared to scenarios with a higher density of vehicles. Similarly, retaining the same number of vehicles while increasing the transmission range leads to extended convergence time compared to the lower transmission range, yet it remains lower than the scenario with fewer vehicles. This is because a higher transmission range results in clusters with a greater number of vehicles but fewer CHs available to collect model parameters. Consequently, with more vehicles participating, it outperforms the scenario with fewer vehicles. However, due to the reduced number of CHs for gathering model parameters and fewer vehicles connecting to the EPC to engage in global model updates, the convergence time increases compared to a lower transmission range.

Figure 6 shows the number of packets transmitted to the EPC for different values of the number of hops and $\alpha$. A setting with a $MAX\_HOP$ count set to zero denotes no clustering, signifying that each vehicle is considered an individual CH responsible for gradient descent and transmission of gradients to the EPC for model parameter updates. As $MAX\_HOP$ increases, the number of CHs decreases. This is attributed to the formation of larger clusters encompassing multiple vehicles within a single cluster, thereby reducing the number of CHs. Moreover, with a higher $\alpha$ value, minor variations in the average relative speed can lead to more frequent state changes. This increased sensitivity can result in more vehicles being identified as CHs or experiencing changes in their roles within the network. This is because the system responds quickly to fluctuations in speed, leading to adjustments in the CH selection process. Consequently, this elevated CH count leads to an increased number of overhead packets dispatched to the EPC. This correspondence between the number of CHs and packet overhead showcases the relationship between the number of clusters and the associated communication costs of the LTE or 5G NR packets. As the number of CHs grows, the packet overhead increases, underscoring the system's reliance on LTE or 5G packets for EPC communication.

\subsection{Performance Comparison of Proposed Algorithm to Benchmark Algorithms}

\begin{figure}[ht]
     \centering
     \begin{subfigure}{0.45\textwidth}
         \centering
         \includegraphics[width=\textwidth]{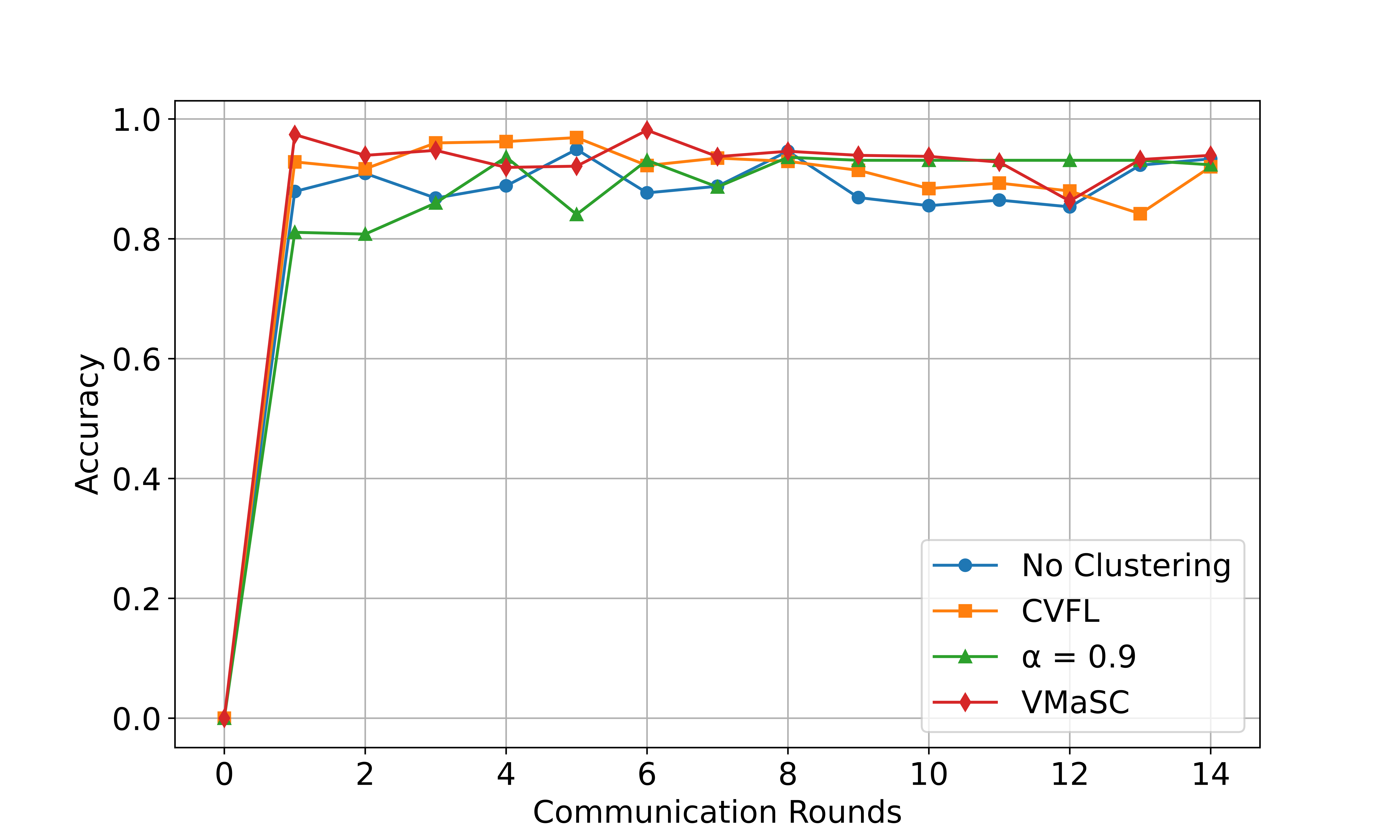}
         \caption{10 Vehicles, Tx Range= 100 Meters}
         \label{fig: 10-100}
     \end{subfigure}
     \vfill
     \begin{subfigure}{0.45\textwidth}
         \centering
         \includegraphics[width=\textwidth]{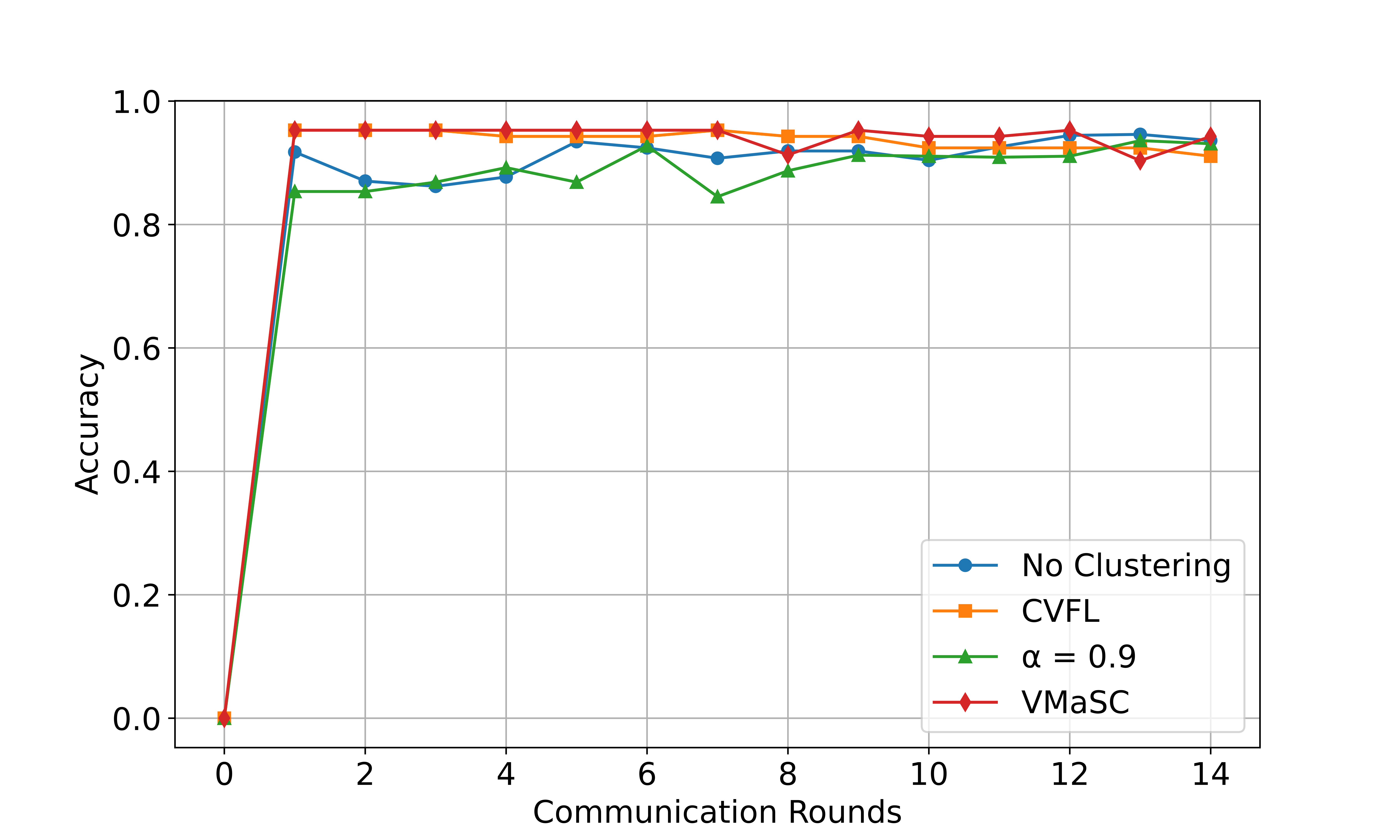}
         \caption{20 Vehicles, Tx Range= 100 Meters}
         \label{fig: 20-100}
     \end{subfigure}
     \vfill
     \begin{subfigure}{0.45\textwidth}
         \centering
         \includegraphics[width=\textwidth]{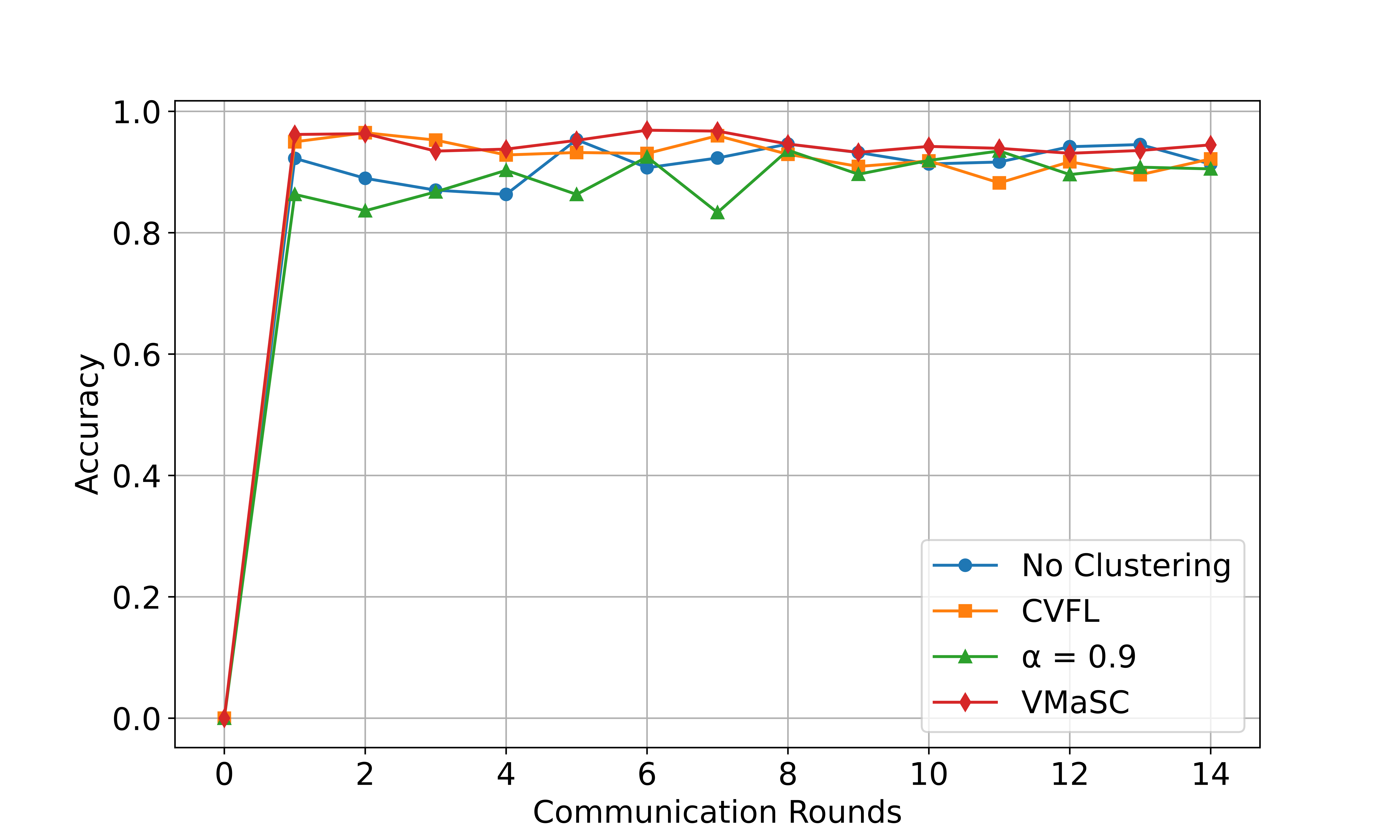}
         \caption{20 Vehicles, Tx Range= 500 Meters}
         \label{fig: 20-500}
     \end{subfigure}
        \caption{Comparison of accuracy in the EPC, using $\alpha=0.9$, with benchmark algorithms for $MAX\_HOP = 1$.}
        \label{fig: EPC comp Accuracy for MAX_HOP=1}
\end{figure}

\begin{figure}[ht]
     \centering
     \begin{subfigure}{0.45\textwidth}
         \centering
         \includegraphics[width=\textwidth]{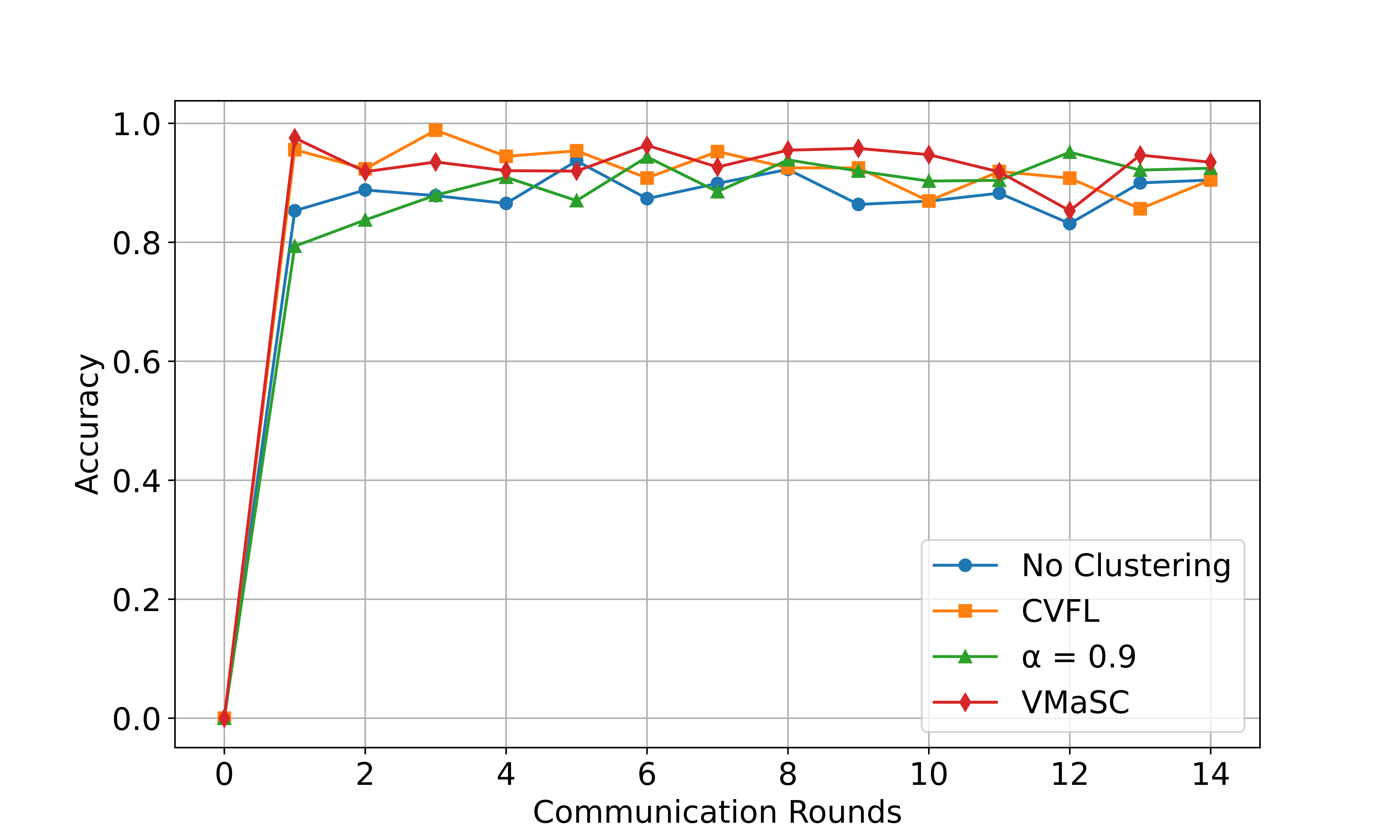}
         \caption{10 Vehicles, Tx Range= 100 Meters}
         \label{fig: 10-100}
     \end{subfigure}
     \vfill
     \begin{subfigure}{0.45\textwidth}
         \centering
         \includegraphics[width=\textwidth]{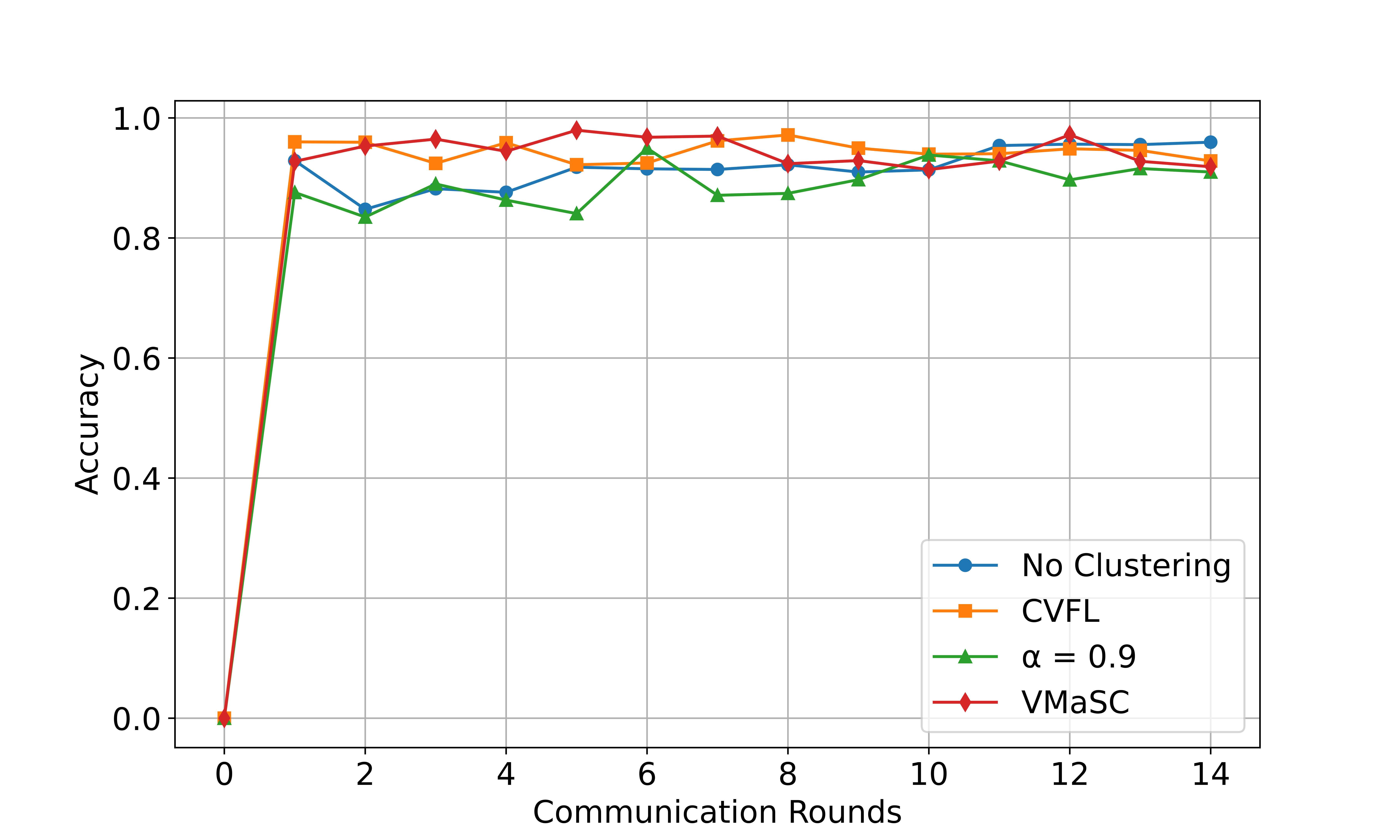}
         \caption{20 Vehicles, Tx Range= 100 Meters}
         \label{fig: 20-100}
     \end{subfigure}
     \vfill
     \begin{subfigure}{0.45\textwidth}
         \centering
         \includegraphics[width=\textwidth]{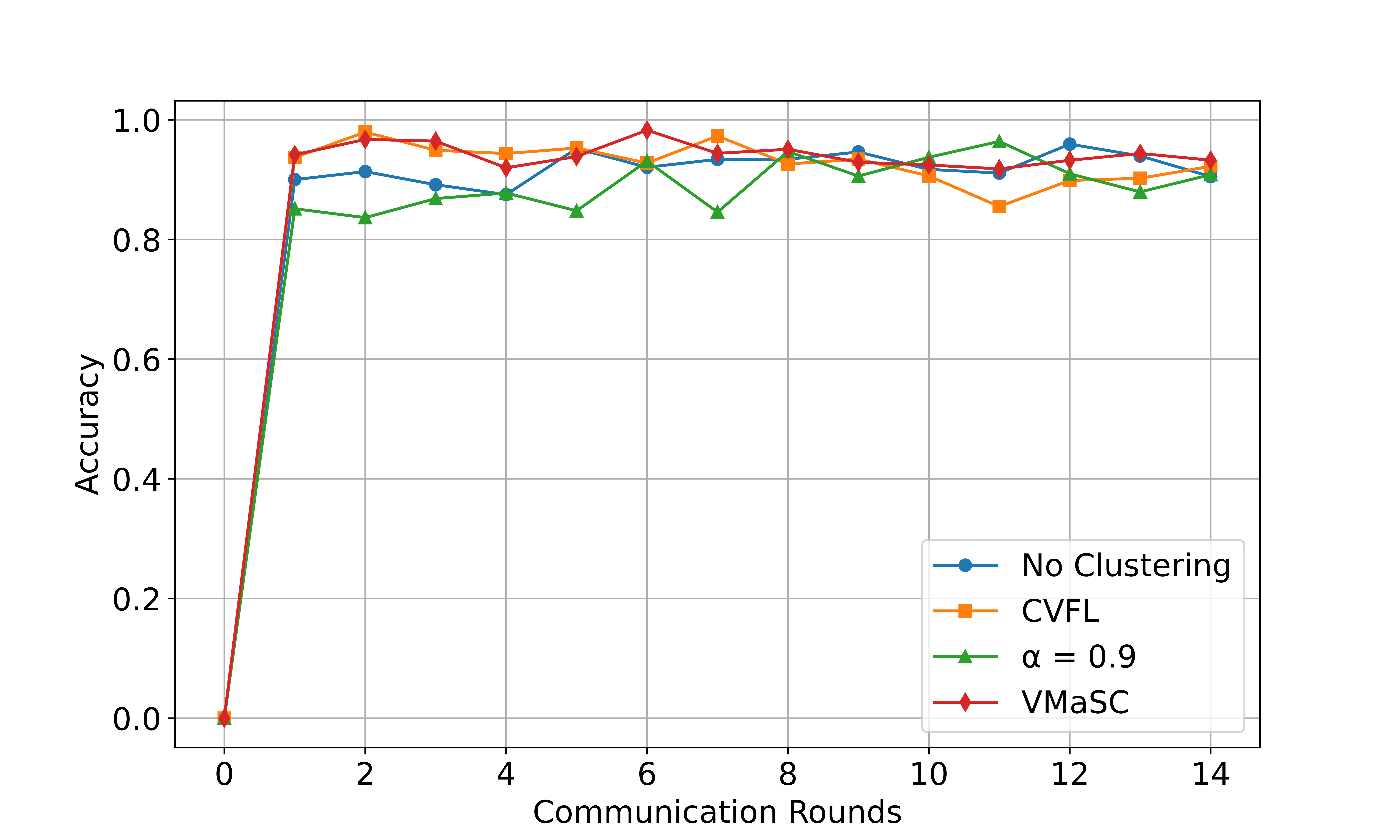}
         \caption{20 Vehicles, Tx Range= 500 Meters}
         \label{fig: 20-500}
     \end{subfigure}
        \caption{Comparison of accuracy in the EPC, using $\alpha=0.9$, with benchmark algorithms for $MAX\_HOP = 2$.}
        \label{fig: EPC comp Accuracy for MAX_HOP=2}
\end{figure}

\begin{figure}[ht]
     \centering
     \begin{subfigure}{0.45\textwidth}
         \centering
         \includegraphics[width=\textwidth]{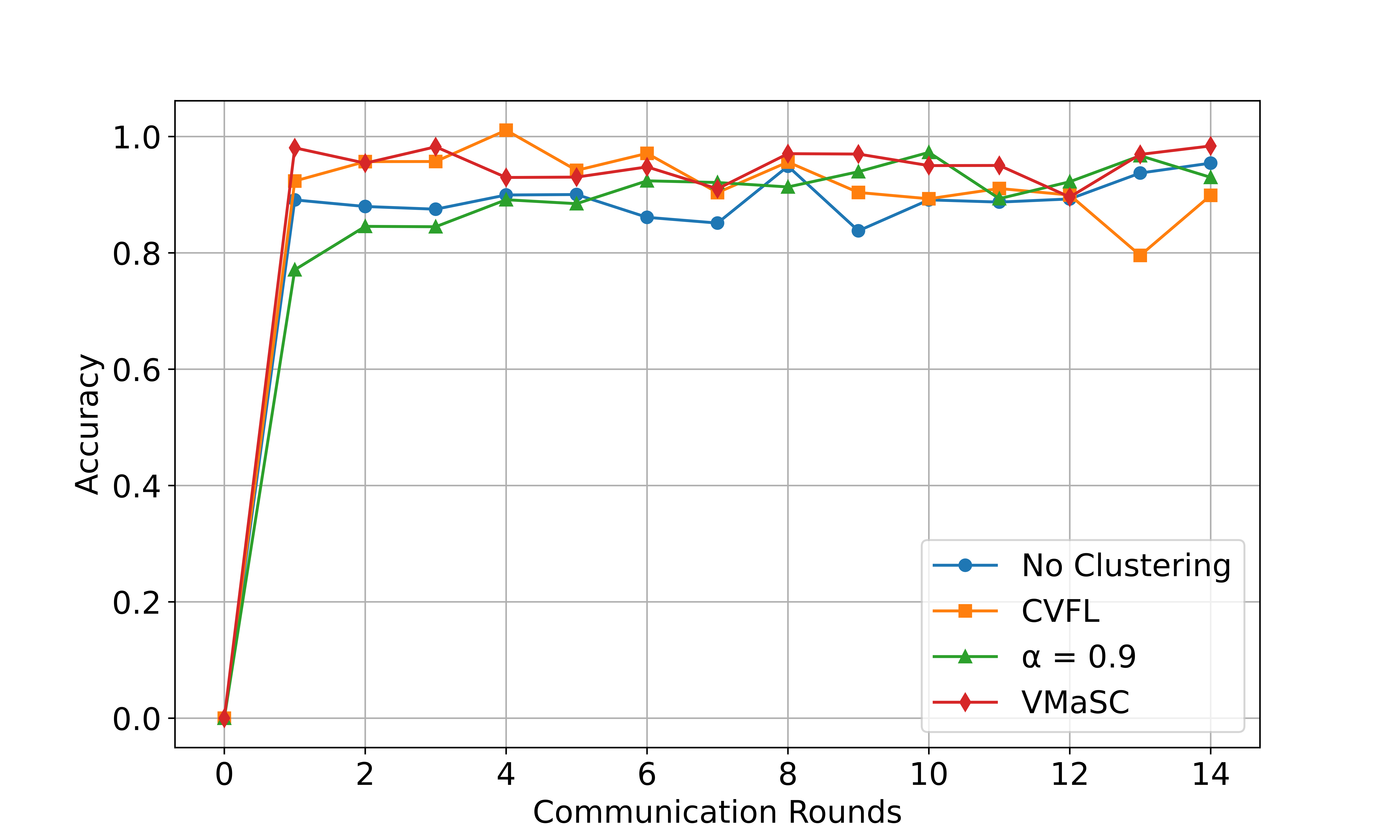}
         \caption{10 Vehicles, Tx Range= 100 Meters}
         \label{fig: 10-100}
     \end{subfigure}
     \vfill
     \begin{subfigure}{0.45\textwidth}
         \centering
         \includegraphics[width=\textwidth]{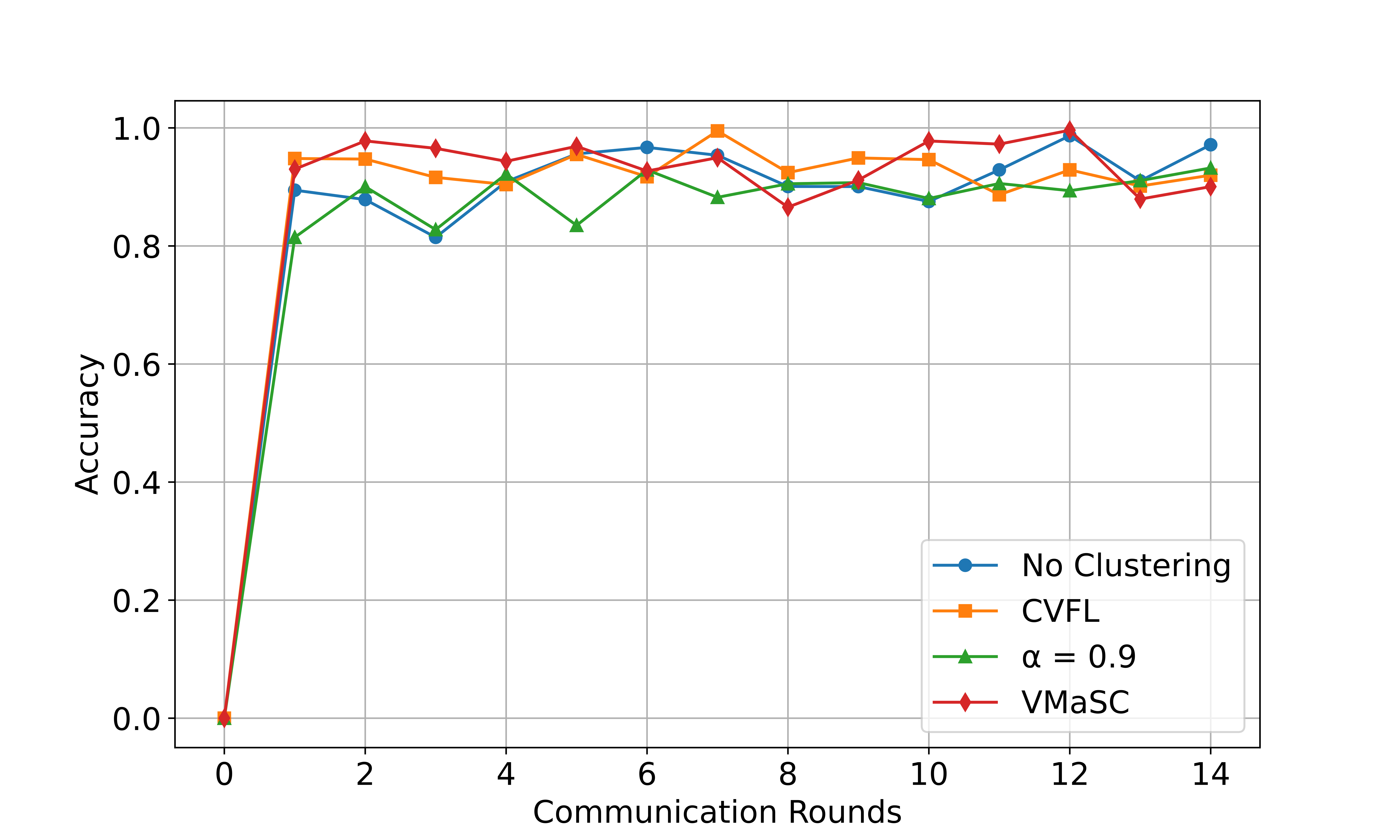}
         \caption{20 Vehicles, Tx Range= 100 Meters}
         \label{fig: 20-100}
     \end{subfigure}
     \vfill
     \begin{subfigure}{0.45\textwidth}
         \centering
         \includegraphics[width=\textwidth]{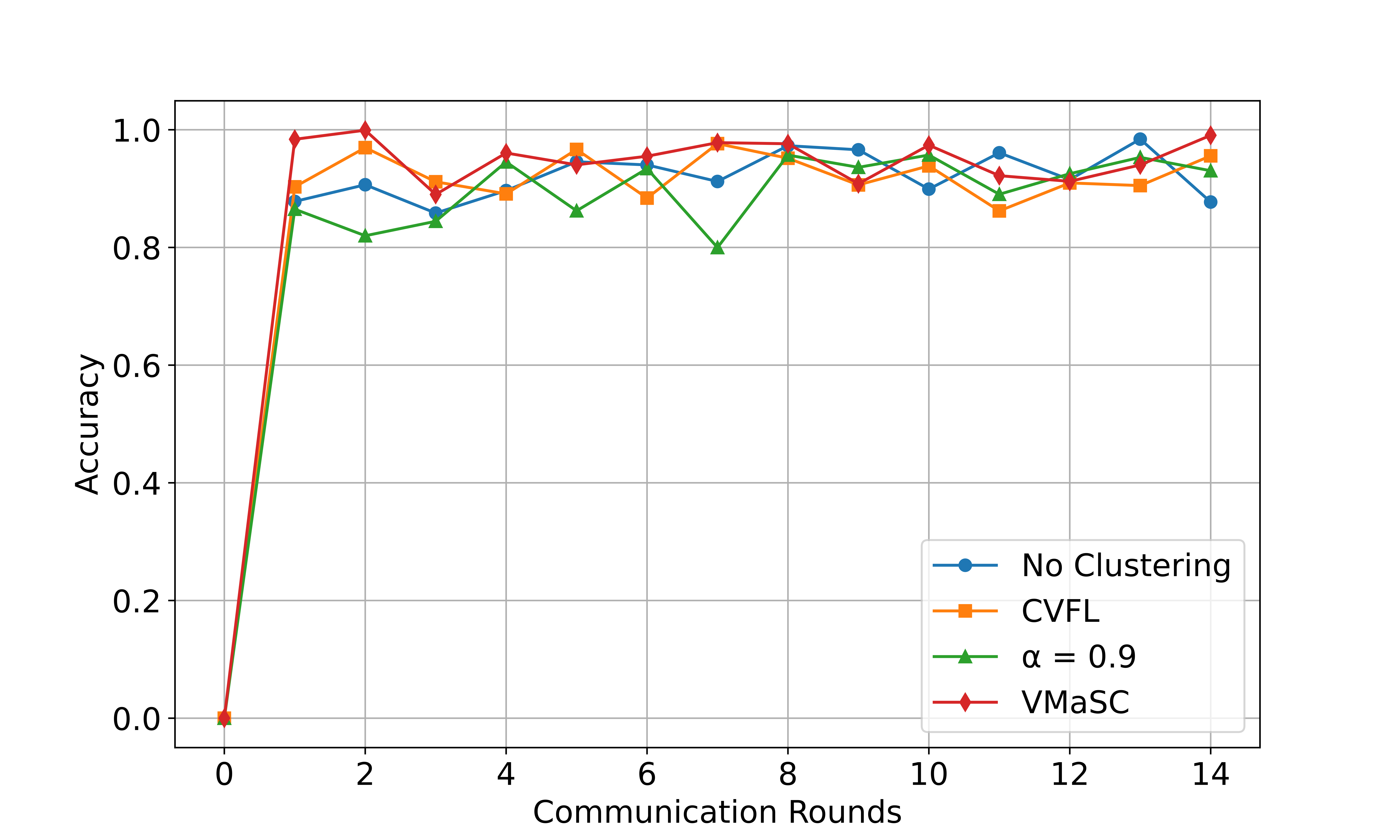}
         \caption{20 Vehicles, Tx Range= 500 Meters}
         \label{fig: 20-500}
     \end{subfigure}
        \caption{Comparison of accuracy in the EPC, using $\alpha=0.9$, with benchmark algorithms for $MAX\_HOP = 3$.}
        \label{fig: EPC comp Accuracy for MAX_HOP=3}
\end{figure}

Figures 7-9 present the EPC accuracy across the same scenarios as Figures 3-5. Our algorithm performance is compared to established benchmarks, including VMaSC, CVFL, and a scenario with no clustering for different values of $MAX\_HOP$. CVFL and VMaSC algorithms exhibit stagnation in local minima, failing to accommodate new data from emerging vehicles or adapting to vehicular and cluster dynamics changes as influenced by the clustering algorithm. In scenarios without clustering, significant fluctuations persist, even in higher communication rounds, indicating a heightened likelihood of divergence. Furthermore, when considering different numbers of hops in the network, it becomes evident that an increase in the number of hops corresponds to an increase in fluctuations. This is primarily attributed to the scenario of fewer clusters with a higher number of participating vehicles in the FL procedure. Consequently, a reduced number of CHs contribute to the model parameter and participate in the acquisition of the global model. Additionally, with an increase in the number of hops, the benchmarks display substantial variability, contrasting with our algorithm, which consistently demonstrates a trend toward higher accuracy with reduced fluctuations. Furthermore, within a fixed number of hops, as the number of vehicles increases under a constant transmission scenario, the degree of fluctuation tends to decrease. Conversely, the benchmark algorithm exhibits a susceptibility to becoming trapped in local minima or experiencing abrupt fluctuations in accuracy during extended communication rounds. In contrast, our algorithm maintains a more stable behavior with reduced fluctuations, offering improved performance and reliability in prolonged communication rounds. On the other hand, within a fixed number of hops and a constant number of vehicles, when the transmission range increases, there is a noticeable increase in the variability observed throughout the FL procedure. This heightened variation is a consequence of having a greater number of vehicles spread across fewer clusters. This phenomenon underscores the challenge faced by benchmark algorithms, which struggle to consistently achieve a high level of accuracy with stable performance under these conditions. In contrast, our algorithm demonstrates a remarkable ability to maintain a stable behavior while achieving high accuracy when compared to other methods. This capability positions our approach as a more robust and reliable solution, particularly in scenarios with a higher transmission range.

\section{Conclusion}

This paper presents a novel framework for HFL over a multi-hop cluster-based vehicular network with the goal of addressing the issues of limited communication resources at high vehicle density, high vehicle mobility, and the statistical diversity of data distributions. The proposed algorithm features a dynamic multi-hop clustering algorithm, combining average relative speed and cosine similarity metrics to ensure stable clustering and fast convergence in HFL. Additionally, the introduced framework incorporates an innovative mechanism aimed at enabling smooth transitions between cluster heads (CHs) and facilitating the transfer of the most up-to-date FL model parameter. This mechanism effectively aligns the FL procedure with the clustering algorithm. Through simulations, we have demonstrated that the proposed algorithm improves accuracy and convergence time while maintaining an acceptable level of packet overhead. This framework provides a promising solution for FL in VANETs, enhancing the efficiency and effectiveness of data-driven decision-making, real-time traffic prediction, and other advanced services.
In the future, we plan to work on analyzing the effect of HFL on security, incorporating vehicle selection into the HFL, and investigating the applicability of the proposed framework in traffic prediction and management, collision avoidance, autonomous driving, environmental monitoring, and emergency services applications.

\newpage

\bibliographystyle{ieeetr}
\bibliography{bare_jrnl}

\end{document}